# Complex Question Answering: Unsupervised Learning Approaches and Experiments

**Yllias Chali**　　　　　　　　　　　　　　　　　　　CHALI@CS.ULETH.CA
*University of Lethbridge*
*Lethbridge, AB, Canada, T1K 3M4*

**Shafiq R. Joty**　　　　　　　　　　　　　　　　　　RJOTY@CS.UBC.CA
*University of British Columbia*
*Vancouver, BC, Canada, V6T 1Z4*

**Sadid A. Hasan**　　　　　　　　　　　　　　　　　　HASAN@CS.ULETH.CA
*University of Lethbridge*
*Lethbridge, AB, Canada, T1K 3M4*

## Abstract

Complex questions that require inferencing and synthesizing information from multiple documents can be seen as a kind of topic-oriented, informative multi-document summarization where the goal is to produce a single text as a compressed version of a set of documents with a minimum loss of relevant information. In this paper, we experiment with one empirical method and two unsupervised statistical machine learning techniques: K-means and Expectation Maximization (EM), for computing relative importance of the sentences. We compare the results of these approaches. Our experiments show that the empirical approach outperforms the other two techniques and EM performs better than K-means. However, the performance of these approaches depends entirely on the feature set used and the weighting of these features. In order to measure the importance and relevance to the user query we extract different kinds of features (i.e. lexical, lexical semantic, cosine similarity, basic element, tree kernel based syntactic and shallow-semantic) for each of the document sentences. We use a local search technique to learn the weights of the features. To the best of our knowledge, no study has used tree kernel functions to encode syntactic/semantic information for more complex tasks such as computing the relatedness between the query sentences and the document sentences in order to generate query-focused summaries (or answers to complex questions). For each of our methods of generating summaries (i.e. empirical, K-means and EM) we show the effects of syntactic and shallow-semantic features over the bag-of-words (BOW) features.

## 1. Introduction

The vast increase in the amount of online text available and the demand for access to different types of information have led to a renewed interest in a broad range of Information Retrieval (IR) related areas that go beyond the simple document retrieval. These areas include question answering, topic detection and tracking, summarization, multimedia retrieval, chemical and biological informatics, text structuring, text mining, genomics, etc. Automated Question Answering (QA)—the ability of a machine to answer questions, simple or complex, posed in ordinary human language—is perhaps the most exciting technological development of the past six or seven years (Strzalkowski & Harabagiu, 2008). The expectations are already tremendous, reaching beyond the discipline (a subfield of Natural Language Processing (NLP)) itself.

As a tool for finding documents on the web, search engines are proven to be adequate. Although there is no limitation in the expressiveness of the user in terms of query formulation, certain limitations exist in what the search engine does with the query. Complex question answering tasks require multi-document summarization through an aggregated search, or a faceted search, that represents an information need which cannot be answered by a single document. For example, if we look for the comparison of the average number of years between marriage and first birth for women in the U.S., Asia, and Europe, the answer is likely contained in multiple documents. Multi-document summarization is useful for this type of query and there is currently no tool on the market that is designed to meet this kind of information need.

QA research attempts to deal with a wide range of question types including: fact, list, definition, how, why, hypothetical, semantically-constrained, and cross-lingual questions. Some questions, which we will call simple questions, are easier to answer. For example, the question: "Who is the president of Bangladesh?" asks for a person's name. This type of question (i.e. factoid) requires small snippets of text as the answer. Again, the question: "Which countries has Pope John Paul II visited?" is a sample of a list question, asking only for a list of small snippets of text.

After having made substantial headway in factoid and list questions, researchers have turned their attention to more complex information needs that cannot be answered by simply extracting named entities (persons, organizations, locations, dates, etc.) from documents. Unlike informationally simple factoid questions, complex questions often seek multiple different types of information simultaneously and do not presuppose that one single answer can meet all of its information needs. For example, with a factoid question like: "How accurate are HIV tests?" it can be safely assumed that the submitter of the question is looking for a number or a range of numbers. However, with complex questions like: "What are the causes of AIDS?" the wider focus of this question suggests that the submitter may not have a single or well-defined information need and therefore may be amenable to receiving additional supporting information that is relevant to some (as yet) undefined informational goal (Harabagiu, Lacatusu, & Hickl, 2006). These questions require inferencing and synthesizing information from multiple documents.

A well known QA systems is the Korean Naver's Knowledge iN search[1], who were the pioneers in community QA. This tool allows users to ask just about any question and get answers from other users. Naver's Knowledge iN now has roughly 10 times more entries than Wikipedia. It is used by millions of Korean web users on any given day. Some people say Koreans are not addicted to the internet but to Naver. As of January 2008 the Knowledge Search database included more than 80 million pages of user-generated information. Another popular answer service is Yahoo! Answers which is a community-driven knowledge market website launched by Yahoo!. It allows users to both submit questions to be answered and answer questions from other users. People vote on the best answer. The site gives members the chance to earn points as a way to encourage participation and is based on the Naver model. As of December 2006, Yahoo! Answers had 60 million users and 65 million answers. Google had a QA system[2] based on paid editors which was launched in April 2002 and fully closed in December 2006.

---

1. http://kin.naver.com/

However, from a computational linguistics point of view *information synthesis* can be seen as a kind of topic-oriented informative multi-document summarization. The goal is to produce a single text as a compressed version of a set of documents with a minimum loss of relevant information. Unlike indicative summaries (which help to determine whether a document is relevant to a particular topic), informative summaries must attempt to find answers.

In this paper, we focus on an extractive approach of summarization where a subset of the sentences in the original documents are chosen. This contrasts with abstractive summarization where the information in the text is rephrased. Although summaries produced by humans are typically not extractive, most of the state of the art summarization systems are based on extraction and they achieve better results than the automated abstraction. Here, we experimented with one empirical and two well-known unsupervised statistical machine learning techniques: K-means and EM and evaluated their performance in generating topic-oriented summaries. However, the performance of these approaches depends entirely on the feature set used and the weighting of these features. In order to measure the importance and relevance to the user query we extract different kinds of features (i.e. lexical, lexical semantic, cosine similarity, basic element, tree kernel based syntactic and shallow-semantic) for each of the document sentences. We have used a gradient descent local search technique to learn the weights of the features.

Traditionally, information extraction techniques are based on the BOW approach augmented by language modeling. But when the task requires the use of more complex semantics, the approaches based on only BOW are often inadequate to perform fine-level textual analysis. Some improvements on BOW are given by the use of dependency trees and syntactic parse trees (Hirao, , Suzuki, Isozaki, & Maeda, 2004; Punyakanok, Roth, & Yih, 2004; Zhang & Lee, 2003b), but these too are not adequate when dealing with complex questions whose answers are expressed by long and articulated sentences or even paragraphs. Shallow semantic representations, bearing more compact information, could prevent the sparseness of deep structural approaches and the weakness of BOW models (Moschitti, Quarteroni, Basili, & Manandhar, 2007). As pinpointing the answer to a question relies on a deep understanding of the semantics of both, attempting an application of syntactic and semantic information to complex QA seems natural. To the best of our knowledge, no study has used tree kernel functions to encode syntactic/semantic information for more complex tasks such as computing the relatedness between the query sentences and the document sentences in order to generate query-focused summaries (or answers to complex questions). For all of our methods of generating summaries (i.e. empirical, K-means and EM) we show the effects of syntactic and shallow-semantic features over the BOW features.

Over the past three years, complex questions have been the focus of much attention in both the automatic question-answering and Multi Document Summarization (MDS) communities. Typically, most current complex QA evaluations including the 2004 AQUAINT Relationship QA Pilot, the 2005 Text Retrieval Conference (TREC) Relationship QA Task, and the TREC definition (and others) require systems to return unstructured lists of can-didate answers in response to a complex question. However recently, MDS evaluations (including the 2005, 2006 and 2007 Document Understanding Conference (DUC)) have tasked systems with returning paragraph-length answers to complex questions that are responsive, relevant, and coherent.

---

2. http://answers.google.com/

Our experiments based on the DUC 2007 data show that including syntactic and semantic features improves the performance. Comparison among the approaches are also shown. Comparing with DUC 2007 participants, our systems achieve top scores and there is no statistically significant difference between the results of our system and the results of DUC 2007 best system.

This paper is organized as follows: Section 2 focuses on the related work, Section 3 gives a brief description of our intended final model, Section 4 describes how the features are extracted, Section 5 discusses the learning issues and presents our learning approaches, Section 6 discusses how we remove the redundant sentences before adding them to the final summary, and Section 7 describes our experimental study. We conclude and discuss future directions in Section 8.

## 2. Related Work

Researchers all over the world working on query-based summarization are trying different directions to see which methods provide the best results.

There are a number of sentence retrieval systems based on IR (Information Retrieval) techniques. These systems typically don't use a lot of linguistic information, but they still deserve special attention. Murdock and Croft (2005) propose a translation model specifically for monolingual data, and show that it significantly improves sentence retrieval over query likelihood. Translation models train on a parallel corpus and they used a corpus of question/answer pairs. Losada (2005) presents a comparison between multiple-Bernoulli models and multinomial models in the context of a sentence retrieval task and shows that a multivariate Bernoulli model can really outperform popular multinomial models for retrieving relevant sentences. Losada and Fernández (2007) propose a novel sentence retrieval method based on extracting highly frequent terms from top retrieved documents. Their results reinforce the idea that top retrieved data is a valuable source to enhance retrieval systems. This is specially true for short queries because there are usually few query-sentence matching terms. They argue that this method improves significantly the precision at top ranks when handling poorly specified information needs.

The *LexRank* method addressed by Erkan and Radev (2004) was very successful in generic multi-document summarization. A topic-sensitive LexRank is proposed by Otterbacher, Erkan, and Radev (2005). As in LexRank, the set of sentences in a document cluster is represented as a graph where nodes are sentences, and links between the nodes are induced by a similarity relation between the sentences. The system then ranks the sentences according to a random walk model defined in terms of both the inter-sentence similarities and the similarities of the sentences to the topic description or question.

Concepts of coherence and cohesion enable us to capture the theme of the text. Coherence represents the overall structure of a multi-sentence text in terms of macro-level relations between clauses or sentences (Halliday & Hasan, 1976). Cohesion, as defined by Halliday and Hasan (1976), is the property of holding text together as one single grammat-ical unit based on relations (i.e. ellipsis, conjunction, substitution, reference, and lexical cohesion) between various elements of the text. Lexical cohesion is defined as the cohesion that arises from the semantic relations (collocation, repetition, synonym, hypernym, hyponym, holonym, meronym, etc.) between the words in the text (Morris & Hirst, 1991). Lexical cohesion among words are represented by lexical chains which are the sequences of semantically related words. The summarization methods based on lexical chain first extract the nouns, compound nouns and named entities as candidate words (Li, Sun, Kit, & Webster, 2007). Then using WordNet[3] the systems find the semantic similarity between the nouns and compound nouns. After that lexical chains are built in two steps:

1. Building single document strong chains while disambiguating the senses of the words.

2. Building a multi-chain by merging the strongest chains of the single documents into one chain.

The systems rank sentences using a formula that involves a) the lexical chain, b) keywords from the query and c) named entities. For example, Li et al. (2007) uses the following formula:

$$Score = \alpha P(chain) + \beta P(query) + \gamma P(namedEntity)$$

where $P(chain)$ is the sum of the scores of the chains whose words come from the candidate sentence, $P(query)$ is the sum of the co-occurrences of key words in a topic and the sentence, and $P(namedEntity)$ is the number of name entities existing in both the topic and the sentence. The three coefficients $\alpha$, $\beta$ and $\gamma$ are set empirically. The top ranked sentences are then selected to form the summary.

Harabagiu et al. (2006) introduce a new paradigm for processing complex questions that relies on a combination of (a) question decompositions; (b) factoid QA techniques; and (c) Multi-Document Summarization (MDS) techniques. The question decomposition procedure operates on a Markov chain. That is, by following a random walk with a mixture model on a bipartite graph of relations established between concepts related to the topic of a complex question and subquestions derived from topic-relevant passages that manifest these relations. Decomposed questions are then submitted to a state-of-the-art QA system in order to retrieve a set of passages that can later be merged into a comprehensive answer by a MDS system. They show that question decompositions using this method can significantly enhance the relevance and comprehensiveness of summary-length answers to complex questions.

There are approaches that are based on probabilistic models (Pingali, K., & Varma, 2007; Toutanova, Brockett, Gamon, Jagarlamudi, Suzuki, & Vanderwende, 2007). Pingali et al. (2007) rank the sentences based on a mixture model where each component of the model is a statistical model:

$$Score(s) = \alpha \times QIScore(s) + (1 - \alpha) \times QFocus(s, Q) \tag{1}$$

---

3. WordNet (http://wordnet.princeton.edu/) is a widely used semantic lexicon for the English language. It groups English words (i.e. nouns, verbs, adjectives and adverbs) into sets of synonyms called synsets, provides short, general definitions (i.e. gloss definition), and records the various semantic relations between these synonym sets.

Where Score(s) is the score for sentence $s$. Query-independent score (QIScore) and query-dependent score (QFocus) are calculated based on probabilistic models. Toutanova et al. (2007) learns a log-linear sentence ranking model by maximizing three metrics of sentence goodness: (a) ROUGE oracle, (b) Pyramid-derived, and (c) Model Frequency. The scoring function is learned by fitting weights for a set of feature functions of sentences in the document set and is trained to optimize a sentence pair-wise ranking criterion. The scoring function is further adapted to apply to summaries rather than sentences and to take into account redundancy among sentences.

Pingali et al. (2007) reduce the document-sentences by dropping words that do not contain any important information. Toutanova et al. (2007), Vanderwende, Suzuki, and Brockett (2006), and Zajic, Lin, Dorr, and Schwartz (2006) heuristically decompose the document-sentences into smaller units. They apply a small set of heuristics to a parse tree to create alternatives after which both the original sentence and (possibly multiple) simplified versions are available for selection.

There are approaches in multi-document summarization that do try to cluster sentences together. Guo and Stylios (2003) use verb arguments (i.e. subjects, times, locations and actions) for clustering. For each sentence this method establishes the indices information based on the verb arguments (subject is first index, time is second, location is third and action is fourth). All the sentences that have the same or closest 'subjects' index are put in a cluster and they are sorted out according to the temporal sequence from the earliest to the latest. Sentences that have the same 'spaces/locations' index value in the cluster are then marked out. The clusters are ranked based on their sizes and top 10 clusters are chosen. Then, applying a cluster reduction module the system generates the compressed extract summaries.

There are approaches in "Recognizing Textual Entailment", "Sentence Alignment", and "Question Answering" that use syntactic and/or semantic information in order to measure the similarity between two textual units. This indeed motivated us to include syntactic and semantic features to get the structural similarity between a document sentence and a query sentence (discussed in Section 4.1). MacCartney, Grenager, de Marneffe, Cer, and Manning (2006) use typed dependency graphs (same as dependency trees) to represent the text and the hypothesis. They try to find a good partial alignment between the typed dependency graphs representing the hypothesis (contains $n$ nodes) and the text (graph contains $m$ nodes) in a search space of $O((m + 1)n)$. They use an incremental beam search combined with a node ordering heuristic to do approximate global search in the space of possible alignments. A locally decomposable scoring function was chosen such that the score of an alignment is the sum of the local node and edge alignment scores. The scoring measure is designed to favor alignments which align semantically similar subgraphs, irrespective of polarity. For this reason, nodes receive high alignment scores when the words they represent are semantically similar. Synonyms and antonyms receive the highest score and unrelated words receive the lowest. Alignment scores also incorporate local edge scores which are based on the shape of the paths between nodes in the text graph which correspond to adjacent nodes in the hypothesis graph. In the final step they make a decision about whether or not the hypothesis is entailed by the text conditioned on the typed dependency graphs as well as the best alignment between them. To make this decision they use a supervised statistical logistic regression classifier (with a feature space of 28 features) with a Gaussian prior parameter for regularization.

Hirao et al. (2004) represent the sentences using Dependency Tree Path (DTP) to incorporate syntactic information. They apply String Subsequence Kernel (SSK) to measure the similarity between the DTPs of two sentences. They also introduce Extended String Subsequence Kernel (ESK) to incorporate semantics in DTPs. Kouylekov and Magnini (2005) use the tree edit distance algorithms on the dependency trees of the text and the hypothesis to recognize the textual entailment. According to this approach, a text $T$ entails a hypothesis $H$ if there exists a sequence of transformations (i.e. deletion, insertion and substitution) applied to $T$ such that we can obtain $H$ with an overall cost below a certain threshold. Punyakanok et al. (2004) represent the question and the sentence containing answer with their dependency trees. They add semantic information (i.e. named entity, synonyms and other related words) in the dependency trees. They apply the approximate tree matching in order to decide how similar any given pair of trees are. They also use the edit distance as the matching criteria in the approximate tree matching. All these methods show the improvement over the BOW scoring methods.

## 3. Our Approach

To accomplish the task of answering complex questions we extract various important features for each of the sentences in the document collection to measure its relevance to the query. The sentences in the document collection are analyzed in various levels and each of the document sentences is represented as a vector of feature-values. Our feature set includes lexical, lexical semantic, statistical similarity, syntactic and semantic features, and graph-based similarity measures (Chali & Joty, 2008b). We reimplemented many of these features which are successfully applied to many related fields of NLP.

We use a simple local search technique to fine-tune the feature weights. We also use the statistical clustering algorithms: EM and K-means to select the relevant sentences for summary generation. Experimental results show that our systems perform better when we include the tree kernel based syntactic and semantic features though summaries based on only syntactic or semantic feature do not achieve good results. Graph-based cosine similarity and lexical semantic features are also important for selecting relevant sentences. We find that the local search technique outperforms the other two and the EM performs better than the K-means based learning. In the later sections we describe all the subparts of our systems in details.

## 4. Feature Extraction

In this section, we will describe the features that will be used to score the sentences. We provide detailed examples[4] to show how we get the feature values. We will first describe the syntactic and semantic features that we are introducing in this work. We follow with a detailed description of the features more commonly used in the question answering and summarization communities.

---

4. All the query and document sentences used in the examples are taken from the DUC 2007 collection.

## 4.1 Syntactic and Shallow Semantic Features

For the task like *query-based summarization* that requires the use of more complex syntactic and semantics, the approaches with only BOW are often inadequate to perform fine-level textual analysis. The importance of syntactic and semantic features in this context is described by Zhang and Lee (2003a), Moschitti et al. (2007), Bloehdorn and Moschitti (2007a), Moschitti and Basili (2006) and Bloehdorn and Moschitti (2007b).

An effective way to integrate syntactic and semantic structures in machine learning algorithms is the use of *tree kernel* functions (Collins & Duffy, 2001; Moschitti & Quarteroni, 2008) which has been successfully applied to question classification (Zhang & Lee, 2003a; Moschitti & Basili, 2006). Syntactic and semantic information are used effectively to measure the similarity between two textual units by MacCartney et al. (2006). To the best of our knowledge, no study has used tree kernel functions to encode syntactic/semantic information for more complex tasks such as computing the relatedness between the query sentences and the document sentences. Another good way to encode some shallow syntactic information is the use of Basic Elements (BE) (Hovy, Lin, Zhou, & Fukumoto, 2006) which uses dependency relations. Our experiments show that including syntactic and semantic features improves the performance on the sentence selection for complex question answering task (Chali & Joty, 2008a).

### 4.1.1 Encoding Syntactic Structures

**Basic Element (BE) Overlap Measure** Shallow syntactic information based on dependency relations was proved to be effective in finding similarity between two textual units (Hirao et al., 2004). We incorporate this information by using Basic Elements that are defined as follows (Hovy et al., 2006):

- The head of a major syntactic constituent (noun, verb, adjective or adverbial phrases), expressed as a single item.

- A relation between a head-BE and a single dependent, expressed as a triple: (head|modifier|relation).

The triples encode some syntactic information and one can decide whether any two units match or not- more easily than with longer units (Hovy et al., 2006). We extracted BEs for the sentences (or query) by using the BE package distributed by ISI[5].

Once we get the BEs for a sentence, we computed the Likelihood Ratio (LR) for each BE following Zhou, Lin, and Hovy (2005). Sorting BEs according to their LR scores produced a BE-ranked list. Our goal is to generate a summary that will answer the users' questions. The ranked list of BEs in this way contains important BEs at the top which may or may not be relevant to the users' questions. We filter those BEs by checking whether they contain any word which is a *query word* or a *QueryRelatedWords* (defined in Section 4.3). For example, if we consider the following sentence we get the BE score of 0.77314.

**Query:** Describe steps taken and worldwide reaction prior to introduction of the Euro on January 1, 1999. Include predictions and expectations reported in the press.

---

5. BE website:http://www.isi.edu/ cyl/BE

**Sentence:** The Frankfurt-based body said in its annual report released today that it has decided on two themes for the new currency: history of European civilization and abstract or concrete paintings.

**BE Score:** 0.77314

Here, the BE "decided|themes|obj" is not considered as it does not contain any word from the query words or query relevant words but BE "report|annual|mod" is taken as it contains a query word *"report"*. In this way, we filter out the BEs that are not related to the query. The score of a sentence is the sum of its BE scores divided by the number of BEs in the sentence. By limiting the number of the top BEs that contribute to the calculation of the sentence scores we can remove the BEs with little importance and the sentences with fewer important BEs. If we set the threshold to 100 only the topmost 100 BEs in the ranked list can contribute to the normalized sentence BE score computation. In this paper, we did not set any threshold— we took all the BEs counted when calculating the BE scores for the sentences.

**Tree Kernels Approach** In order to calculate the syntactic similarity between the query and the sentence we first parse the sentence as well as the query into a syntactic tree (Moschitti, 2006) using a parser like Charniak (1999). Then we calculate the similarity between the two trees using the *tree kernel*. We reimplemented the tree kernel model as proposed by Moschitti et al. (2007).

Once we build the trees, our next task is to measure the similarity between the trees. For this, every tree $T$ is represented by an $m$ dimensional vector $v(T) = (v_1(T), v_2(T), \cdots v_m(T))$, where the i-th element $v_i(T)$ is the number of occurrences of the i-th tree fragment in tree $T$. The tree fragments of a tree are all of its sub-trees which include at least one production with the restriction that no production rules can be broken into incomplete parts (Moschitti et al., 2007) Figure 1 shows an example tree and a portion of its subtrees.

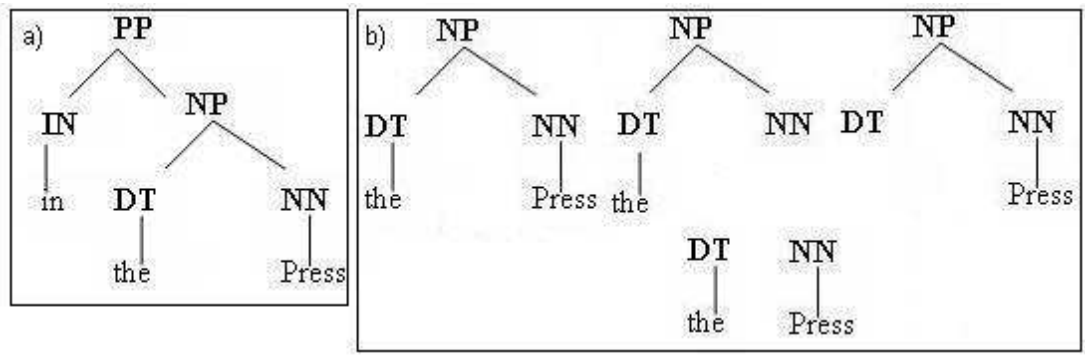

Figure 1: (a) An example tree (b) The sub-trees of the NP covering "the press".

Implicitly we enumerate all the possible tree fragments $1, 2, \cdots, m$. These fragments are the axis of this m-dimensional space. Note that this could be done only implicitly since the number $m$ is extremely large. Because of this, Collins and Duffy (2001) define the tree kernel algorithm whose computational complexity does not depend on $m$.

The tree kernel of two trees $T_1$ and $T_2$ is actually the inner product of $v(T_1)$ and $v(T_2)$:

$$TK(T_1, T_2) = v(T_1).v(T_2) \tag{2}$$

We define the indicator function $I_i(n)$ to be 1 if the sub-tree $i$ is seen rooted at node $n$ and 0 otherwise. It follows:

$$v_i(T_1) = \sum_{n_1 \in N_1} I_i(n_1), v_i(T_2) = \sum_{n_2 \in N_2} I_i(n_2) \tag{3}$$

Where $N_1$ and $N_2$ are the set of nodes in $T_1$ and $T_2$ respectively. So, we can derive:

$$
\begin{aligned}
TK(T_1, T_2) & = v(T_1).v(T_2) = \sum_i v_i(T_1)v_i(T_2) \\
& = \sum_{n_1 \in N_1} \sum_{n_2 \in N_2} \sum_i I_i(n_1)I_i(n_2) \\
& = \sum_{n_1 \in N_1} \sum_{n_2 \in N_2} C(n_1, n_2)
\end{aligned}
\tag{4}
$$

where we define $C(n_1, n_2) = \sum_i I_i(n_1)I_i(n_2)$. Next, we note that $C(n_1, n_2)$ can be computed in polynomial time due to the following recursive definition:

1. If the productions at $n_1$ and $n_2$ are different then $C(n_1, n_2) = 0$

2. If the productions at $n_1$ and $n_2$ are the same, and $n_1$ and $n_2$ are pre-terminals, then $C(n_1, n_2) = 1$

3. Else if the productions at $n_1$ and $n_2$ are not pre-terminals,

$$C(n_1, n_2) = \prod_{j=1}^{nc(n_1)} (1 + C(ch(n_1, j), ch(n_2, j))) \tag{5}$$

where $nc(n_1)$ is the number of children of $n_1$ in the tree; because the productions at $n_1$ and $n_2$ are the same we have $nc(n_1) = nc(n_2)$. The i-th child-node of $n_1$ is $ch(n_1, i)$.

In cases where the query is composed of two or more sentences we compute the similarity between the document sentence ($s$) and each of the query-sentences ($q_i$) then we take the average of the scores as the syntactic feature value.

$$Syntactic\ similarity\ value = \frac{\sum_{i=1}^n TK(q_i, s)}{n}$$

Where $n$ is the number of sentences in the query $q$ and $s$ is the sentence under consideration. $TK$ is the similarity value (tree kernel) between the sentence $s$ and the query sentence $q$ based on the syntactic structure. For example, for the following sentence $s$ and query $q$ we get the score:

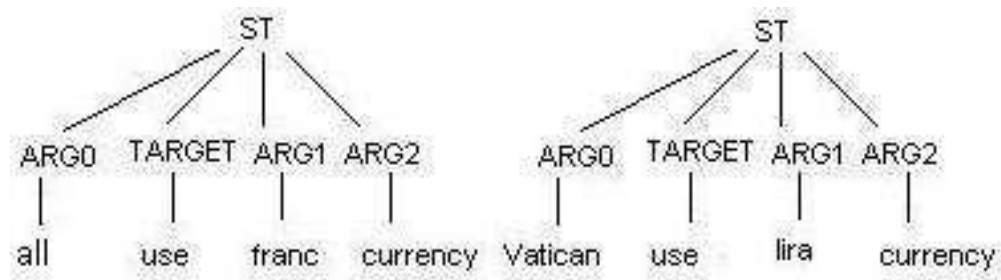

Figure 2: Example of semantic trees

**Query (q):** Describe steps taken and worldwide reaction prior to introduction of the Euro on January 1, 1999. Include predictions and expectations reported in the press.

**Sentence (s):** Europe's new currency, the euro, will rival the U.S. dollar as an international currency over the long term, Der Spiegel magazine reported Sunday.

**Scores:** 90, 41

**Average Score:** 65.5

### 4.1.2 SEMANTIC FEATURES

Though introducing syntactic information gives an improvement on BOW, by the use of syntactic parses, this too is not adequate when dealing with complex questions whose answers are expressed by long and articulated sentences or even paragraphs. Shallow semantic representations, bearing more compact information, could prevent the sparseness of deep structural approaches and the weakness of BOW models (MacCartney et al., 2006; Moschitti et al., 2007).

Initiatives such as PropBank (PB) (Kingsbury & Palmer, 2002) have made the design of accurate automatic Semantic Role Labeling (SRL) systems like ASSERT (Hacioglu, Pradhan, Ward, Martin, & Jurafsky, 2003) possible. Hence, attempting an application of SRL to QA seems natural as pinpointing the answer to a question relies on a deep understanding of the semantics of both. For example, consider the PB annotation:

[ARG0 all][TARGET use][ARG1 the french franc][ARG2 as their currency]

Such annotation can be used to design a shallow semantic representation that can be matched against other semantically similar sentences, e.g.

[ARG0 the Vatican][TARGET use][ARG1 the Italian lira][ARG2 as their currency]

In order to calculate the semantic similarity between the sentences we first represent the annotated sentence (or query) using the tree structures like Figure 2 called Semantic Tree (ST) as proposed by Moschitti et al. (2007). In the semantic tree arguments are replaced with the most important word–often referred to as the semantic head. We look for a noun first, then a verb, then an adjective, then adverb to find the semantic head in the argument. If none of these is present we take the first word of the argument as the semantic head.

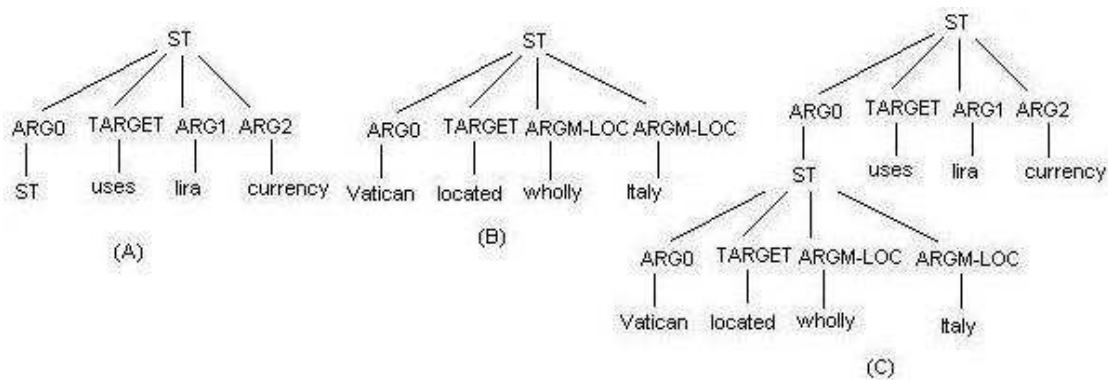

Figure 3: Two STs composing a STN

However, sentences rarely contain a single predicate, rather typically propositions contain one or more subordinate clauses. For instance, let us consider a slight modification of the second sentence: "the Vatican, located wholly within Italy uses the Italian lira as their currency." Here, the main predicate is "uses" and the subordinate predicate is "located". The SRL system outputs the following two annotations:

(1) [ARG0 the Vatican located wholly within Italy][TARGET uses][ARG1 the Italian lira][ARG2 as their currency]

(2) [ARG0 the Vatican][TARGET located] [ARGM-LOC wholly][ARGM-LOC within Italy] uses the Italian lira as their currency

giving the STs in Figure 3. As we can see in Figure 3(A), when an argument node corresponds to an entire subordinate clause we label its leaf with ST (e.g. the leaf of ARG0). Such ST node is actually the root of the subordinate clause in Figure 3(B). If taken separately, such STs do not express the whole meaning of the sentence. Hence, it is more accurate to define a single structure encoding the dependency between the two predicates as in Figure 3(C). We refer to this kind of nested STs as STNs.

Note that the tree kernel (TK) function defined in Section 4.1.1 computes the number of common subtrees between two trees. Such subtrees are subject to the constraint that their nodes are taken with all or none of the children they have in the original tree. Though this definition of subtrees makes the TK function appropriate for syntactic trees, it is not well suited for the semantic trees (ST). For instance, although the two STs of Figure 2 share most of the subtrees rooted in the $ST$ node, the kernel defined above computes no match.

The critical aspect of steps (1), (2), and (3) of the TK function is that the productions of two evaluated nodes have to be identical to allow the match of further descendants. This means that common substructures cannot be composed by a node with only some of its children as an effective ST representation would require. Moschitti et al. (2007) solve this problem by designing the Shallow Semantic Tree Kernel (SSTK) which allows portions of an ST to match.

**Shallow Semantic Tree Kernel (SSTK)**   We reimplemented the SSTK according to the model given by Moschitti et al. (2007). The SSTK is based on two ideas: first, it changes the ST, as shown in Figure 4 by adding $SLOT$ nodes. These accommodate argument labels in a specific order with a fixed number of slots, possibly filled with $null$ arguments that encode all possible predicate arguments. Leaf nodes are filled with the wildcard character * but they may alternatively accommodate additional information. The slot nodes are used in such a way that the adopted TK function can generate fragments containing one or more children like for example those shown in frames (b) and (c) of Figure 4. As previously pointed out, if the arguments were directly attached to the root node the kernel function would only generate the structure with all children (or the structure with no children, i.e. empty) (Moschitti et al., 2007).

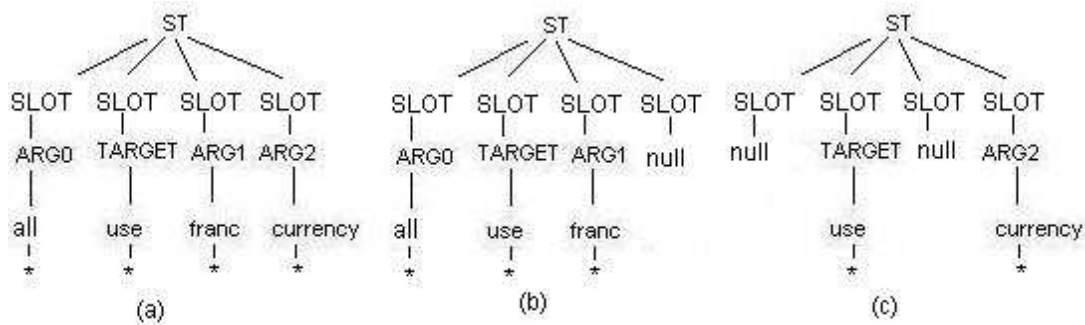

Figure 4: Semantic tree with some of its fragments

Second, as the original tree kernel would generate many matches with slots filled with the null label we have set a new step 0 in the TK calculation:

(0) if $n_1$ (or $n_2$) is a pre-terminal node and its child label is $null$, $C(n_1, n_2) = 0$;
    and subtract one unit to $C(n_1, n_2)$, in step 3:

$$(3)\ C(n_1, n_2) = \prod_{j=1}^{nc(n_1)} \left(1 + C(ch(n_1, j), ch(n_2, j))\right) - 1 \qquad (6)$$

The above changes generate a new $C$ which, when substituted (in place of original $C$) in Eq. 4, gives the new SSTK.

For example, for the following sentence $s$ and query $q$ we get the semantic score:

**Query (q):** Describe steps taken and worldwide reaction prior to introduction of the Euro on January 1, 1999. Include predictions and expectations reported in the press.

**Sentence (s):** The Frankfurt-based body said in its annual report released today that it has decided on two themes for the new currency history of European civilization and abstract or concrete paintings.

**Scores:** 6, 12

**Average Score:** 9

## 4.2 Lexical Features

Here, we will discuss the lexical features that are most commonly used in the QA and summarization communities. We reimplemented all of them in this research.

### 4.2.1 N-GRAM OVERLAP

N-gram overlap measures the overlapping word sequences between the candidate document sentence and the query sentence. With the view to measure the overlap scores, a *query pool* and a *sentence pool* are created. In order to create the query (or sentence) pool, we took the query (or document) sentence and created a set of related sentences by replacing its content words[6] by their first-sense synonyms using WordNet. For example, given a stemmed document-sentence: "John write a poem", the sentence pool contains: "John compose a poem", "John write a verse form" along with the given sentence.

We measured the recall based n-gram scores for a sentence $P$ using the following formula:

$$NgramScore(P) = max_i(max_j \ Ngram(s_i, q_j)) \tag{7}$$

$$Ngram(S, Q) = \frac{\sum_{gram_n \in S} Count_{match}(gram_n)}{\sum_{gram_n \in S} Count(gram_n)} \tag{8}$$

Where $n$ stands for the length of the n-gram ($n = 1, 2, 3, 4$), and $Count_{match}(gram_n)$ is the number of n-grams co-occurring in the query and the candidate sentence, $q_j$ is the j-th sentence in the query pool, and $s_i$ is the i-th sentence in the sentence pool of sentence $P$.

**1-gram Overlap Measure**

A 1-gram overlap score measures the number of words common in the sentence in hand and the *query related words*. This can be computed as follows:

$$1gram \ Overlap \ Score = \frac{\sum_{w_1 \in S} Count_{match}(w_1)}{\sum_{w_1 \in S} Count(w_1)} \tag{9}$$

Where $S$ is the set of content words in the candidate sentence and $Count_{match}$ is the number of matches between the *sentence content words* and *query related words*. $Count(gram_n)$ is the number of $w_1$.

Note that in order to measure the 1-gram score we took the *query related words* instead of the exact query words. The motivation behind this is the sentence which has word(s) that are not exactly the query words but their synonyms, hypernyms, hyponym or gloss words, will get counted.

**Example:**

**Query** Describe steps taken and worldwide reaction prior to introduction of the Euro on January 1, 1999. Include predictions and expectations reported in the press.

**Sentence** The Frankfurt-based body said in its annual *study* released today that it has decided on two themes for the new currency: history of European civilization and abstract or concrete paintings.

---

6. hence forth content words are the nouns, verbs, adverbs and adjectives.

**1-gram Score** 0.06666 (After normalization[7]).

Note that the above sentence has a 1-gram overlap score of 0.06666 even though it has no exact word common with the query words. It got this score because the sentence word *study* is a synonym of the query word *report*.

**Other N-gram Overlap Measures**

As above, we can calculate the other n-gram overlap scores. For example, considering the following query sentence and document sentence (From DUC 2007 collection), we have 4 matching 2-grams: ("1 1999", "of Euro", "on January" and "January 1"). Hence, employing the formula given above, we get the following 2-gram score after normalization. 3-gram score is also found accordingly.

**Query Sentence:** Describe steps taken and worldwide reaction prior to introduction of the Euro on January 1, 1999. Include predictions and expectations reported in the press.

**Document Sentence:** Despite skepticism about the actual realization of a single European currency as scheduled on *January 1, 1999*, preparations for the design *of the Euro* note have already begun.

**2-gram:** 0.14815

**3-gram:** 0.0800

### 4.2.2 LCS AND WLCS

A sequence $W = [w_1, w_2, ..., w_n]$ is a subsequence of another sequence $X = [x_1, x_2, ..., x_m]$, if there exists a strict increasing sequence $[i_1, i_2, ..., i_n]$ of indices of X such that for all $j = 1, 2, ..., n$ we have $x_{i_j} = w_j$ (Cormen, Leiserson, & Rivest, 1989). Given two sequences $S_1$ *and* $S_2$, the longest common subsequence (LCS) of $S_1$ *and* $S_2$ is a common subsequence with maximum length (Lin, 2004).

The longer the LCS of two sentences is, the more similar the two sentences are. We used LCS-based F-measure to estimate the similarity between the document sentence $S$ of length $m$ and the query sentence $Q$ of length $n$ as follows:

$$R_{lcs}(S, Q) = \frac{LCS(S, Q)}{m} \tag{10}$$

$$P_{lcs}(S, Q) = \frac{LCS(S, Q)}{n} \tag{11}$$

$$F_{lcs}(S, Q) = (1 - \alpha) \times P_{lcs}(S, Q) + \alpha \times R_{lcs}(S, Q) \tag{12}$$

Where $LCS(S, Q)$ is the length of a longest common subsequence of $S$ and $Q$, and $\alpha$ is a constant that determines the importance of precision and recall. While computing the LCS measure each document sentence and query sentence are viewed as a sequence of words.

---

7. We normalize each of the feature values corresponding to a sentence with respect to the entire context of a particular document.

The intuition is that the longer the LCS of these two is the more similar they are. Here the recall ($R_{lcs}(S,Q)$) is the ratio of the *length of the longest common subsequence of S and Q* to the *document sentence length* that measures the completeness. Whereas the precision ($P_{lcs}(S,Q)$) is the ratio of the *length of the longest common subsequence of S and Q to the query sentence length* which is a measure of exactness. To obtain the equal importance to precision and recall we set the value of $\alpha$ as 0.5. Equation 12 is called the LCS-based F-measure. Notice that $F_{lcs}$ is 1 when, S=Q; and $F_{lcs}$ is 0 when there is nothing in common between $S$ and $Q$.

One advantage of using LCS is that it does not require consecutive matches but in-sequence matches that reflect sentence level word order as n-grams. The other advantage is that it automatically includes longest in-sequence common n-grams. Therefore, no predefined n-gram length is necessary. Moreover, it has the property that its value is less than or equal to the minimum of the unigram (i.e. 1-gram) F-measure of S and Q. Unigram recall reflects the proportion of words in S that are also present in Q; while unigram precision is the proportion of words in Q that are also in S. Unigram recall and precision count all co-occurring words regardless of their orders; while LCS counts in-sequence co-occurrences.

By only awarding credit to in-sequence unigram matches, LCS measure also captures sentence level structure in a natural way. Consider the following example:

**S1** *John shot the thief*

**S2** John shot the thief

**S3** the thief shot John

Using S1 as reference sentence, and S2 and S3 as the sentences under consideration S2 and S3 would have the same 2-gram score since they both have one bigram (i.e. "the thief") in common with S1. However, S2 and S3 have very different meanings. In case of LCS S2 has a score of 3/4=0.75 and S3 has a score of 2/4=0.5 with $\alpha$ = 0.5. Therefore, S2 is better than S3 according to LCS.

However, LCS suffers one disadvantage in that it only counts the main in-sequence words; therefore, other alternative LCSes and shorter sequences are not reflected in the final score. For example, given the following candidate sentence:

**S4** the thief John shot

Using S1 as its reference, LCS counts either "the thief" or "John shot" but not both; therefore, S4 has the same LCS score as S3 while 2-gram would prefer S4 over S3.

In order to measure the LCS score for a sentence we took a similar approach as the previous section using WordNet (i.e. creation of sentence pool and query pool). We calculated the LCS score using the following formula:

$$LCS\ score = max_i(max_j\ F_{lcs}(s_i, q_j)) \tag{13}$$

Where $q_j$ is the j-th sentence in the query pool, and $s_i$ is the i-th sentence in the sentence pool.

The basic LCS has a problem in that it does not differentiate LCSes of different spatial relations within their embedding sequences (Lin, 2004). For example, given a reference sequence **S** and two candidate sequences $Y_1$ *and* $Y_2$ as follows:

**S:** <u>A</u> <u>B</u> <u>C</u> <u>D</u> E F G

$Y_1$ : <u>A</u> <u>B</u> <u>C</u> <u>D</u> H I K

$Y_2$ : <u>A</u> H <u>B</u> K <u>C</u> I <u>D</u>

$Y_1$ *and* $Y_2$ have the same LCS score. However, $Y_1$ should be better choice than $Y_2$ because $Y_1$ has consecutive matches. To improve the basic LCS method we can store the length of consecutive matches encountered so far to a regular two dimensional dynamic program table computing LCS. We call it weighted LCS (WLCS) and use k to indicate the length of the current consecutive matches ending at words $x_i$ and $y_j$. Given two sentences X and Y, the WLCS score of X and Y can be computed using the similar dynamic programming procedure as stated by Lin (2004). We use WLCS as it has the advantage of not measuring the similarity by taking the words in a higher dimension like string kernels which indeed reduces the time complexity. As before, we computed the WLCS-based F-measure in the same way using both the query pool and the sentence pool.

$$WLCS\ score = max_i(max_j\ F_{wlcs}(s_i, q_j)) \tag{14}$$

**Example:**

**Query Sentence:** Describe steps taken and worldwide reaction prior to introduction of the Euro on January 1, 1999. Include predictions and expectations reported in the press.

**Document Sentence:** Despite skepticism about the actual realization of a single European currency as scheduled on *January 1, 1999*, preparations for the design *of the Euro* note have already begun.

We find 6 matching strings: ("of on 1 Euro 1999 January") in the longest common subsequence considering this sentence and related sentences. For WLCS we set the weight as 1.2. After normalization, we get the following LCS and WLCS scores for the sentence applying the above formula.

**LCS Score:** 0.27586

**WLCS Score:** 0.15961

### 4.2.3 Skip-Bigram Measure

A skip-bigram is any pair of words in their sentence order allowing for arbitrary gaps. Skip-bigram measures the overlap of skip-bigrams between a candidate sentence and a query sentence (Lin, 2004). We rely on the query pool and the sentence pool as before using WordNet. Considering the following sentences:

**S1** *John shot the thief*

**S2** John shoot the thief

**S3** the thief shoot John

**S4** the thief John shot

we get that each sentence has C(4,2)=6 skip-bigrams[8]. For example, S1 has the following skip-bigrams: ("John shot", "John the", "John thief", "shot the", "shot thief" and "the thief") S2 has three skip bi-gram matches with S1 ("John the", "John thief", "the thief"), S3 has one skip bi-gram match with S1 ("the thief"), and S4 has two skip bi-gram matches with S1 ("John shot", "the thief").

The skip bi-gram score between the document sentence $S$ of length $m$ and the query sentence $Q$ of length $n$ can be computed as follows:

$$R_{skip_2}(S, Q) = \frac{SKIP_2(S, Q)}{C(m, 2)} \qquad (15)$$

$$P_{skip_2}(S, Q) = \frac{SKIP_2(S, Q)}{C(n, 2)} \qquad (16)$$

$$F_{skip_2}(S, Q) = (1 - \alpha) \times P_{skip_2}(S, Q) + \alpha \times R_{skip_2}(S, Q) \qquad (17)$$

Where $SKIP_2(S, Q)$ is the number of skip bi-gram matches between $S$ and $Q$, and $\alpha$ is a constant that determines the importance of precision and recall. We set the value of $\alpha$ as 0.5 to associate the equal importance to precision and recall. $C$ is the combination function. We call the equation 17 the skip bigram-based F-measure. We computed the skip bigram-based F-measure using the formula:

$$SKIP\ BIGRAM = max_i(max_j\ F_{skip2}(s_i, q_j)) \qquad (18)$$

For example, given the following query and the sentence, we get 8 skip-bigrams: ("on 1", "January 1", "January 1999", "of Euro", "1 1999", "on 1999", "on January" and "of on"). Applying the equations above, we get skip bi-gram score of 0.05218 after normalization.

**Query** Describe steps taken and worldwide reaction prior to introduction of the Euro on January 1, 1999. Include predictions and expectations reported in the press.

**Sentence** Despite skepticism about the actual realization of a single European currency as scheduled on *January 1, 1999*, preparations for the design *of the Euro* note have already begun.

**Skip bi-gram Score:** 0.05218

---

8. $C(n, r) = \frac{n!}{r! \times (n-r)!}$

Note that skip bi-gram counts all in-order matching word pairs while LCS only counts one longest common subsequence. We can put the constraint on the maximum skip distance, $d_{skip}$, between two in-order words to form a skip bi-gram which avoids the spurious matches like "the the" or "of from". For example, if we set $d_{skip}$ to 0 then it is equivalent to bi-gram overlap measure (Lin, 2004). If we set $d_{skip}$ to 4 then only word pairs of at most 4 words apart can form skip bi-grams. In our experiment we set $d_{skip} = 4$ in order to ponder at most 4 words apart to get the skip bi-grams.

Modifying the equations: 15, 16, and 17 to allow the maximum skip distance limit is straightforward: following Lin (2004) we count the skip bi-gram matches, $SKIP_2(S, Q)$, within the maximum skip distance and replace the denominators of the equations with the actual numbers of within distance skip bi-grams from the reference sentence and the candidate sentence respectively.

### 4.2.4 Head and Head Related-words Overlap

The number of head words common in between two sentences can indicate how much they are relevant to each other. In order to extract the heads from the sentence (or query), the sentence (or query) is parsed by Minipar[9] and from the dependency tree we extract the heads which we call exact head words. For example, the head word of the sentence: "John eats rice" is "eat".

We take the synonyms, hyponyms, and hypernyms[10] of both the query-head words and the sentence-head words and form a set of words which we call head-related words. We measured the exact head score and the head-related score as follows:

$$ExactHeadScore = \frac{\sum_{w_1 \in HeadSet} Count_{match}(w_1)}{\sum_{w_1 \in HeadSet} Count(w_1)} \qquad (19)$$

$$HeadRelatedScore = \frac{\sum_{w_1 \in HeadRelSet} Count_{match}(w_1)}{\sum_{w_1 \in HeadRelSet} Count(w_1)} \qquad (20)$$

Where $HeadSet$ is the set of head words in the sentence and $Count_{match}$ is the number of matches between the $HeadSet$ of the query and the sentence. $HeadRelSet$ is the set of synonyms, hyponyms, and hypernyms of head words in the sentence and $Count_{match}$ is the number of matches between the head-related words of the query and the sentence. For example, below we list the head words for a query and a sentence and their measures:

**Query:** Describe steps taken and worldwide reaction prior to introduction of the Euro on January 1, 1999. Include predictions and expectations reported in the press.

**Heads for Query:** include, reaction, step, take, describe, report, Euro, introduction, press, prediction, 1999, expectation

**Sentence:** The Frankfurt-based body said in its annual report released today that it has decided on two themes for the new currency: history of European civilization and abstract or concrete paintings.

---

9. http://www.cs.ualberta.ca/ lindek/minipar.htm
10. hypernym and hyponym levels are restricted to 2 and 3 respectively.

**Heads for Sentence:** history, release, currency, body, report,painting, say, civilization, theme, decide.

**Exact Head Score:** $\frac{1}{11} = 0.09$

**Head Related Score:** 0

### 4.3 Lexical Semantic Features

We form a set of words which we call *QueryRelatedWords* by taking the content words from the query, their first-sense synonyms, the nouns' hypernyms/hyponyms, and the nouns' gloss definitions using WordNet.

#### 4.3.1 SYNONYM OVERLAP

The synonym overlap measure is the overlap between the list of synonyms of the content words extracted from the candidate sentence and *query related words*. This can be computed as follows:

$$Synonym\ Overlap\ Score = \frac{\sum_{w_1 \in SynSet} Count_{match}(w_1)}{\sum_{w_1 \in SynSet} Count(w_1)} \tag{21}$$

Where SynSet is the synonym set of the content words in the sentence and $Count_{match}$ is the number of matches between the SynSet and *query related words*.

#### 4.3.2 HYPERNYM/HYPONYM OVERLAP

The hypernym/hyponym overlap measure is the overlap between the list of hypernyms (level 2) and hyponyms (level 3) of the nouns extracted from the sentence in consideration and *query related words*. This can be computed as follows:

$$Hypernym/hyponym\ overlap\ score = \frac{\sum_{h_1 \in HypSet} Count_{match}(h_1)}{\sum_{h_1 \in HypSet} Count(h_1)} \tag{22}$$

Where HypSet is the hyponym/hyponym set of the nouns in the sentence and $Count_{match}$ is the number of matches between the HypSet and *query related words*.

#### 4.3.3 GLOSS OVERLAP

The gloss overlap measure is the overlap between the list of content words that are extracted from the gloss definition of the nouns in the sentence in consideration and *query related words*. This can be computed as follows:

$$Gloss\ Overlap\ Score = \frac{\sum_{g_1 \in GlossSet} Count_{match}(g_1)}{\sum_{g_1 \in GlossSet} Count(g_1)} \tag{23}$$

Where GlossSet is the set of content words (i.e. nouns, verbs and adjectives) taken from the gloss definition of the nouns in the sentence and $Count_{match}$ is the number of matches between the GlossSet and query related words.

**Example:**

For example, given the query the following sentence gets synonym overlap score of 0.33333, hypernym/hyponym overlap score of 0.1860465 and gloss overlap score of 0.1359223.

**Query** Describe steps taken and worldwide reaction prior to introduction of the Euro on January 1, 1999. Include predictions and expectations reported in the press.

**Sentence** The Frankfurt-based body said in its annual report released today that it has decided on two themes for the new currency: history of European civilization and abstract or concrete paintings.

**Synonym Overlap Score:** 0.33333

**Hypernym/Hyponym Overlap Score:** 0.1860465

**Gloss Overlap Score:** 0.1359223

### 4.4 Statistical Similarity Measures

Statistical similarity measures are based on the co-occurrence of similar words in a corpus. Two words are termed as similar if they belong to the same context. We used the thesaurus provided by Dr. Dekang Lin[11] for these purpose. We have used two statistical similarity measures:

**Dependency-based similarity measure**

This method uses the dependency relations among words in order to measure the similarity (Lin, 1998b). It extracts the dependency triples and then uses a statistical approach to measure the similarity. Using the given corpus one can retrieve the most similar words for a given word. The similar words are grouped into clusters.

Note that for a word there can be more than one cluster. Each cluster represents the sense of the word and its similar words for that sense. So, selecting the right cluster for a word is itself a problem. Our goals are: i) to create a bag of similar words to the query words and ii) once we get the bag of similar words (dependency based) for the query words to measure the overlap score between it and the sentence words.

**Creating Bag of Similar Words:**

For each query-word we extract all of its clusters from the thesaurus. Now in order to determine the right cluster for a query word we measure the overlap score between the *query related words* (i.e. exact words, synonyms, hypernyms/hyponyms and gloss) and the *clusters*. The hypothesis is that the cluster that has more words in common with the query related words is the right cluster under the assumption that the first synonym is the correct sense. We choose the cluster for a word which has the highest overlap score.

$$Overlap\ score_i = \frac{\sum_{w_1 \in QueryRelatedWords} Count_{match}(w_1)}{\sum_{w_1 \in QueryRelatedWords} Count(w_1)} \tag{24}$$

$$Cluster = argmax_i(Overlap\ Score_i) \tag{25}$$

---

11. http://www.cs.ualberta.ca/ lindek/downloads.htm

where QueryRelatedWords is the set of exact words, synonyms, hyponyms/hypernyms, and gloss words for the words in the query (i.e query words) and $Count_{match}$ is the number of matches between the query related words and the $i^{th}$ cluster of similar words.

**Measuring Overlap Score:**

Once we get the clusters for the query words we measured the overlap between the cluster words and the sentence words which we call dependency based similarity measure:

$$DependencyMeasure = \frac{\sum_{w_1 \in SenWords} Count_{match}(w_1)}{\sum_{w_1 \in SenWords} Count(w_1)} \quad (26)$$

Where SenWords is the set of words for the sentence and $Count_{match}$ is the number of matches between the sentence words and the cluster of similar words.

**Proximity-based similarity measure**

This similarity is computed based on the linear proximity relationship between words only (Lin, 1998a). It uses the information theoretic definition of similarity to measure the similarity. The similar words are grouped into clusters. We took the similar approach to measure this feature as the previous section except that we used a different thesaurus.

**Example:**

Considering the following query and sentence we get the following measures:

**Query:** Describe steps taken and worldwide reaction prior to introduction of the Euro on January 1, 1999. Include predictions and expectations reported in the press.

**Sentence:** The Frankfurt-based body said in its annual report released today that it has decided on two themes for the new currency: history of European civilization and abstract or concrete paintings.

**Dependency-based Similarity Score:** 0.0143678

**Proximity-based Similarity Score:** 0.04054054

## 4.5 Graph-based Similarity Measure

Erkan and Radev (2004) used the concept of graph-based centrality to rank a set of sentences for producing generic multi-document summaries. A similarity graph is produced for the sentences in the document collection. In the graph each node represents a sentence. The edges between nodes measure the cosine similarity between the respective pair of sentences. The degree of a given node is an indication of how important the sentence is. Figure 5 shows an example of a similarity graph for 4 sentences.

Once the similarity graph is constructed, the sentences are ranked according to their eigenvector centrality. The LexRank performed well in the context of generic summarization. To apply LexRank to query-focused context a topic-sensitive version of LexRank is proposed by Otterbacher et al. (2005). We followed a similar approach in order to calculate this feature. The score of a sentence is determined by a mixture model of the relevance of the sentence to the query and the similarity of the sentence to other high-scoring sentences.

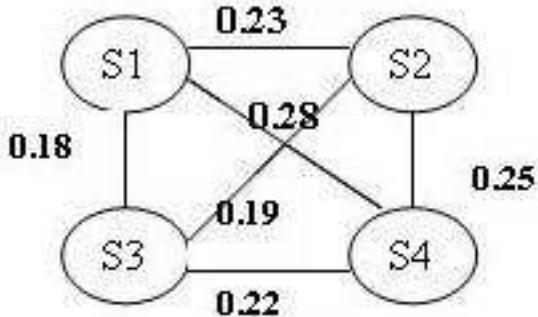

Figure 5: LexRank similarity

## Relevance to the Question

We first stem out all the sentences in the collection and compute the word IDFs (Inverse Document Frequency) using the following formula:

$$idf_w = log\left(\frac{N+1}{0.5 + sf_w}\right) \tag{27}$$

Where $N$ is the total number of sentences in the cluster, and $sf_w$ is the number of sentences that the word $w$ appears in.

We also stem out the questions and remove the stop words. The relevance of a sentence $s$ to the question $q$ is computed by:

$$rel(s|q) = \sum_{w \in q} log\left(tf_{w,s} + 1\right) \times log\left(tf_{w,q} + 1\right) \times idf_w \tag{28}$$

Where $tf_{w,s}$ and $tf_{w,q}$ are the number of times $w$ appears in $s$ and $q$, respectively.

## Mixture Model

In the previous section we measured the relevance of a sentence to the question but a sentence that is similar to the high scoring sentences in the cluster should also have a high score. For instance, if a sentence that gets a high score based on the question relevance model is likely to contain an answer to the question then a related sentence, which may not be similar to the question itself, is also likely to contain an answer (Otterbacher et al., 2005).

We capture this idea by the following mixture model:

$$p(s|q) = \left\{d \times \frac{rel(s|q)}{\sum_{z \in C} rel(z|q)} + (1-d) \times \sum_{v \in C} \frac{sim(s,v)}{\sum_{z \in C} sim(z,v)}\right\} \times p(v|q) \tag{29}$$

Where $p(s|q)$, the score of a sentence $s$ given a question $q$, is determined as the sum of its relevance to the question and the similarity to the other sentences in the collection. $C$ is the set of all sentences in the collection. The value of the parameter $d$ which we call

"bias" is a trade-off between two terms in the equation and is set empirically. For higher values of $d$ we prefer the relevance to the question to the similarity to other sentences. The denominators in both terms are for normalization. Although it is computationally expensive, equation 29 calculates the sum over the entire collection since it is required for the model to sense the global impact through the voting of all sentences. We measure the cosine similarity weighted by word IDFs as the similarity between two sentences in a cluster:

$$sim(x, y) = \frac{\sum_{w \in x, y} tf_{w,x} \times tf_{w,y} \times (idf_w)^2}{\sqrt{\sum_{x_i \in x} (tf_{x_i, x} \times idf_{x_i})^2} \times \sqrt{\sum_{y_i \in y} (tf_{y_i, y} \times idf_{y_i})^2}} \quad (30)$$

Equation 29 can be written in matrix notation as follows:

$$\mathbf{p} = [d\mathbf{A} + (1 - d)\mathbf{B}]^T \mathbf{p} \quad (31)$$

$\mathbf{A}$ is the square matrix such that for a given index $i$, all the elements in the i-th column are proportional to $rel(i|q)$. $\mathbf{B}$ is also a square matrix such that each entry $\mathbf{B}(i,j)$ is proportional to $sim(i,j)$. Both matrices are normalized so that row sums add up to 1. Note that as a result of this normalization all rows of the resulting square matrix $Q = [dA + (1 - d)B]$ also add up to 1. Such a matrix is called *stochastic* and defines a Markov chain. If we view each sentence as a state in a Markov chain then Q(i,j) specifies the transition probability from state $i$ to state $j$ in the corresponding Markov chain. The vector $\mathbf{p}$ we are looking for in Eq. 31 is the stationary distribution of the Markov chain. An intuitive interpretation of the stationary distribution can be understood by the concept of a random walk on the graph representation of the Markov chain. With probability $d$ a transition is made from the current node to the nodes that are similar to the query. With probability *(1-d)* a transition is made to the nodes that are lexically similar to the current node. Every transition is weighted according to the similarity distributions. Each element of the vector $\mathbf{p}$ gives the asymptotic probability of ending up at the corresponding state in the long run regardless of the starting state. The stationary distribution of a Markov chain can be computed by a simple iterative algorithm called *power method* (Erkan & Radev, 2004). It starts with a uniform distribution. At each iteration the eigenvector is updated by multiplying with the transpose of the stochastic matrix. Since the Markov chain is irreducible and aperiodic the algorithm is guaranteed to terminate.

## 5. Ranking Sentences

We use several methods in order to rank sentences to generate summaries applying the features described in Section 4. In this section we will describe the systems in detail.

### 5.1 Learning Feature-weights: A Local Search Strategy

In order to fine-tune the weights of the features, we have used a local search technique. Initially we set all the feature-weights, $w_1, \cdots, w_n$, as equal values (i.e. 0.5) (see Algorithm 1). Then we train the weights using the DUC 2006 data set. Based on the current weights we score the sentences and generate summaries accordingly. We evaluate the summaries using

**Input**: Stepsize $l$, Weight Initial Value $v$
**Output**: A vector $\vec{w}$ of learned weights
Initialize the weight values $w_i$ to $v$.
**for** $i \leftarrow 1\ to\ n$ **do**
    $rg1 = rg2 = prev = 0$
    **while** *(true)* **do**
        scoreSentences($\vec{w}$)
        generateSummaries()
        $rg2 = $ evaluateROUGE()
        **if** $rg1 \leq rg2$ **then**
            prev $= w_i$
            $w_i+ = l$
            $rg1 = rg2$
        **else**
            break
        **end**
    **end**
**end**
return $\vec{w}$

**Algorithm 1**: Tuning weights using Local Search technique

the automatic evaluation tool ROUGE (Lin, 2004) (described in Section 7) and the ROUGE value works as the feedback to our learning loop. Our learning system tries to maximize the ROUGE score in every step by changing the weights individually by a specific step size (i.e. 0.01). That means, to learn weight $w_i$ we change the value of $w_i$ keeping all other weight values ($w_j \forall_{j \neq i}$) stagnant. For each weight $w_i$ the algorithm achieves the local maximum (i.e. hill climbing) of ROUGE value.

Once we have learned the feature-weights we compute the final scores for the sentences using the formula:

$$score_i = \vec{x_i} . \vec{w} \tag{32}$$

Where $\vec{x_i}$ is the feature vector for i-th sentence, $\vec{w}$ is the weight vector, and $score_i$ is the score of i-th sentence.

## 5.2 Statistical Machine Learning Approaches

We experimented with two unsupervised statistical learning techniques with the features extracted in the previous section for the sentence selection problem:

1. K-means learning

2. Expectation Maximization (EM) learning

### 5.2.1 The K-means Learning

*K-means* is a hard clustering algorithm that defines clusters by the center of mass of their members. We start with a set of initial cluster centers that are chosen randomly and go through several iterations of assigning each object to the cluster whose center is closest. After all objects have been assigned we recompute the center of each cluster as the centroid or mean ($\boldsymbol{\mu}$) of its members. The distance function we use is squared Euclidean distance instead of the true Euclidean distance.

Since the square root is a monotonically growing function squared Euclidean distance has the same result as the true Euclidean distance but the computation overload is smaller when the square root is dropped.

Once we have learned the means of the clusters using the K-means algorithm our next task is to rank the sentences according to a probability model. We have used Bayesian model in order to do so. Bayes' law says:

$$
\begin{aligned}
P(q_k|\boldsymbol{x}, \Theta) &= \frac{p(\boldsymbol{x}|q_k, \Theta)P(q_k|\Theta)}{p(\boldsymbol{x}|\Theta)} \\
&= \frac{p(\boldsymbol{x}|q_k, \Theta)P(q_k|\Theta)}{\sum_{k=1}^{K} p(\boldsymbol{x}|q_k, \Theta)p(q_k|\Theta)}
\end{aligned}
\tag{33}
$$

where $q_k$ is a cluster, $\mathbf{x}$ is a feature vector representing a sentence, and $\Theta$ is the parameter set of all class models. We set the weights of the clusters as equiprobable (i.e. $P(q_k|\Theta) = 1/K$). We calculated $p(\boldsymbol{x}|q_k, \Theta)$ using the gaussian probability distribution. The gaussian probability density function (pdf) for the d-dimensional random variable $\boldsymbol{x}$ is given by:

$$
p_{(\boldsymbol{\mu}, \boldsymbol{\Sigma})}(\boldsymbol{x}) = \frac{e^{\frac{-1}{2}(\boldsymbol{x}-\boldsymbol{\mu})^T \boldsymbol{\Sigma}^{-1}(\boldsymbol{x}-\boldsymbol{\mu})}}{\sqrt{2\pi}^d \sqrt{det(\boldsymbol{\Sigma})}}
\tag{34}
$$

where $\boldsymbol{\mu}$, the mean vector, and $\boldsymbol{\Sigma}$, the covariance matrix, are the parameters of the gaussian distribution. We get the means ($\boldsymbol{\mu}$) from the K-means algorithm and we calculate the covariance matrix using the unbiased covariance estimation procedure:

$$
\hat{\boldsymbol{\Sigma}}_j = \frac{1}{N-1} \sum_{i=1}^{N} (\boldsymbol{x}_i - \boldsymbol{\mu}_j)(\boldsymbol{x}_i - \boldsymbol{\mu}_j)^T
\tag{35}
$$

### 5.2.2 THE EM LEARNING

The EM algorithm for gaussian mixture models is a well known method for cluster analysis. A useful outcome of this model is that it produces a likelihood value of the clustering model and the likelihood values can be used to select the best model from a number of different models providing that they have the same number of parameters (i.e. same number of clusters).

**Input**: A sample of $n$ data-points ($\boldsymbol{x}$) each represented by a feature vector of length $L$

**Input**: Number of Clusters $K$

**Output**: An array $\boldsymbol{S}$ of K-means-based Scores

**Data**: Array $\boldsymbol{d}_{nK}$, $\boldsymbol{\mu}_K$, $\boldsymbol{\Sigma}_K$

**Data**: Array $\boldsymbol{C}_K$, $\boldsymbol{y}_{nK}$

Randomly choose $K$ data-points as $K$ initial means: $\boldsymbol{\mu}_k,\ k = 1, \cdots, K$.

**repeat**

    **for** $i \leftarrow 1\ to\ n$ **do**

        **for** $j \leftarrow 1\ to\ K$ **do**

$$\boldsymbol{d}_{ij} \;=\; \|\boldsymbol{x}_i - \boldsymbol{\mu}_j\|^2 = (\boldsymbol{x}_i - \boldsymbol{\mu}_j)^T (\boldsymbol{x}_i - \boldsymbol{\mu}_j)$$

        **end**

        **if** $\boldsymbol{d}_{ik} < \boldsymbol{d}_{il}, \forall l \neq k$ **then**

            assign $\boldsymbol{x}_i$ to $\boldsymbol{C}_k$.

        **end**

    **end**

    **for** $i \leftarrow 1\ to\ K$ **do**

        $\boldsymbol{\mu}_i = \frac{\sum_{\boldsymbol{x}_j \in \boldsymbol{C}_i} \boldsymbol{x}_j}{|\boldsymbol{C}_i|}$

    **end**

**until** *no further change occurs* ;

`/* calculating the covariances for each cluster                    */`

**for** $i \leftarrow 1\ to\ K$ **do**

    $m = |\boldsymbol{C}_i|$

    **for** $j \leftarrow 1\ to\ m$ **do**

        $\boldsymbol{\Sigma}_i +=\ (\boldsymbol{C}_{ij} - \boldsymbol{\mu}_i)\ *\ (\boldsymbol{C}_{ij} - \boldsymbol{\mu}_i)^T$

    **end**

    $\boldsymbol{\Sigma}_i *=\ (1/(m-1))$

**end**

`/* calculating the scores for sentences                            */`

**for** $i \leftarrow 1\ to\ n$ **do**

    **for** $j \leftarrow 1\ to\ K$ **do**

        $y_{ij} = \frac{e^{\frac{-1}{2}(\boldsymbol{x}_i - \boldsymbol{\mu}_j)^T \boldsymbol{\Sigma}_j^{-1}(\boldsymbol{x}_i - \boldsymbol{\mu}_j)}}{\sqrt{2\pi}^d \sqrt{det(\boldsymbol{\Sigma}_j)}}$

    **end**

    **for** $j \leftarrow 1\ to\ K$ **do**

        $z_{ij} = (y_{ij} * w_j)/\sum_{j=1}^{K} y_{ij} * w_j$ ;                  `// where,` $w_j = 1/K$

    **end**

    $m = max(\boldsymbol{\mu}_k)\ \forall k$

    Push $z_{im}\ to\ \boldsymbol{S}$

**end**

return $\boldsymbol{S}$

**Algorithm 2**: Computing K-means based similarity measure

A significant problem with the EM algorithm is that it converges to a local maximum of the likelihood function and hence the quality of the result depends on the initialization. This problem along with a method for improving the initialization is discussed later in this section.

EM is a "soft" version of the K-means algorithm described above. As with K-means we start with a set of random cluster centers $c_1 \cdots c_k$. In each iteration we do a soft assignment of the data-points to every cluster by calculating their membership probabilities. EM is an iterative two step procedure: 1. Expectation-step and 2. Maximization-step. In the expectation step we compute expected values for the hidden variables $h_{i,j}$ which are cluster membership probabilities. Given the current parameters we compute how likely it is that an object belongs to any of the clusters. The maximization step computes the most likely parameters of the model given the cluster membership probabilities.

The data-points are considered to be generated by a mixture model of k-gaussians of the form:

$$P(\mathbf{x}) = \sum_{i=1}^{k} P(C=i)P(\mathbf{x}|C=i) = \sum_{i=1}^{k} P(C=i)P(\mathbf{x}|\boldsymbol{\mu}_i, \boldsymbol{\Sigma}_i) \qquad (36)$$

where the total likelihood of model $\Theta$ with k components, given the observed data points $X = \boldsymbol{x}_1, \cdots, \boldsymbol{x}_n$, is:

$$L(\Theta|X) \quad = \quad \prod_{i=1}^{n} \sum_{j=1}^{k} P(C=j)P(\boldsymbol{x}_i|\Theta_j) = \prod_{i=1}^{n} \sum_{j=1}^{k} w_j P(\boldsymbol{x}_i|\boldsymbol{\mu}_j, \boldsymbol{\Sigma}_j) \qquad (37)$$

$$\Leftrightarrow \quad \sum_{i=1}^{n} log \sum_{j=1}^{k} w_j P(\boldsymbol{x}_i|\boldsymbol{\mu}_j, \boldsymbol{\Sigma}_j) \; ( \; taking \; the \; log \; likelihood \; ) \qquad (38)$$

where $P$ is the probability density function (i.e. eq 34). $\boldsymbol{\mu}_j$ and $\boldsymbol{\Sigma}_j$ are the mean and covariance matrix of component $j$, respectively. Each component contributes a proportion, $w_j$, of the total population such that: $\sum_{j=1}^{K} w_j = 1$.

Log likelihood can be used instead of likelihood as it turns the product into a sum. We describe the EM algorithm for estimating a gaussian mixture.

**Singularities** The covariance matrix $\boldsymbol{\Sigma}$ above must be non-singular or invertible. The EM algorithm may converge to a position where the covariance matrix becomes singular ($|\boldsymbol{\Sigma}| = 0$) or close to singular, that means it is not invertible anymore. If the covariance matrix becomes singular or close to singular then EM may result in wrong clusters. We restrict the covariance matrices to become singular by testing these cases at each iteration of the algorithm as follows:

$$if \; (\sqrt{|\boldsymbol{\Sigma}|} > 1e^{-9}) \; then \; update \; \boldsymbol{\Sigma}$$
$$else \; do \; not \; update \; \boldsymbol{\Sigma}$$

**Discussion: Starting values for the EM algorithm**

The convergence rate and success of clustering using the EM algorithm can be degraded by a poor choice of starting values for the means, covariances, and weights of the components. We experimented with one summary (for document number D0703A from DUC 2007) in order to test the impact of these initial values on the EM algorithm. The cluster means are initialized with a heuristic that spreads them randomly around $Mean(DATA)$ with standard deviation $\sqrt{Cov(DATA) * 10}$. Their initial covariance is set to $Cov(DATA)$ and the initial values of the weights are $w_j = 1/K$ where $K$ is the number of clusters.

That is, for d-dimensional data-points the parameters of $j-th$ component are as follows:

$$\begin{aligned} \vec{\mu_j} &= rand(1, \cdots, d) * \sqrt{\mathbf{\Sigma}(DATA) * 10} + \vec{\mu}(DATA) \\ \mathbf{\Sigma}_j &= \mathbf{\Sigma}(DATA) \\ w_j &= 1/K \end{aligned}$$

The highly variable nature of the results of the tests is reflected in the very inconsistent values for the total log likelihood and the results of repeated experiments indicated that using random starting values for initial estimates of the means frequently gave poor results. There are two possible solutions to this problem. In order to get good results from using random starting values (as specified by the algorithm) we will run the EM algorithm several times and choose the initial configuration for which we get the maximum log likelihood among all configurations. Choosing the best one among several runs is a very computer intensive process. So, to improve the outcome of the EM algorithm on gaussian mixture models, it is necessary to find a better method of estimating initial means for the components.

The best starting position for the EM algorithm, in regard to the estimates of the means, would be to have one estimated mean per cluster which is closer to the true mean of that cluster.

To achieve this aim we explored the widely used "K-means" algorithm as a cluster (means) finding method. That is, the means found by the K-means clustering above will be utilized as the initial means for the EM and we calculate the initial covariance matrices using the unbiased covariance estimation procedure (Equation 35).

**Ranking the Sentences**

Once the sentences are clustered by the EM algorithm, we identify the sentences which are question-relevant by checking their probabilities, $P(q_r|\boldsymbol{x}_i, \Theta)$ where $q_r$ denotes the cluster "question-relevant". If for a sentence $\boldsymbol{x}_i$, $P(q_r|\boldsymbol{x}_i, \Theta) > 0.5$ then $\boldsymbol{x}_i$ is considered to be question-relevant. The cluster which has the mean values greater than the other one is considered as the question-relevant cluster.

Our next task is to rank the question-relevant sentences in order to include them in the summary. This can be done easily by multiplying the feature vector $\vec{x}_i$ with the weight vector $\vec{w}$ that we learned by applying the local search technique (Equation 32).

**Input**: A Sample of $n$ data-points ( $\boldsymbol{x}$) each represented by a feature vector of length
$L$

**Input**: Number of Clusters $K$

**Output**: An array $S$ of EM-based Scores

Start with K initial Gaussian models: $N(\boldsymbol{\mu}_k, \boldsymbol{\Sigma}_k)$ $k = 1, \cdots, K$, with equal priors set
to $P(q_k) = 1/K$.

**repeat**

    `/* Estimation step: compute the probability` $P(q_k^{(i)}|\boldsymbol{x}_j, \Theta^{(i)})$ `for each`

        `data point` $x_j$, $j = 1, \cdots, n$, `to belong to the class` $q_k^{(i)}$         `*/`

    **for** $j \leftarrow 1$ *to* $n$ **do**

        **for** $k \leftarrow 1$ *to* $K$ **do**

$$
\begin{aligned}
P(q_k^{(i)}|\boldsymbol{x}_j, \Theta^{(i)}) &= \frac{P(q_k^{(i)}|\Theta^{(i)})p(\boldsymbol{x}_j|q_k^{(i)}, \Theta^{(i)})}{p(\boldsymbol{x}_j|\Theta^{(i)})} \\
&= \frac{P(q_k^{(i)}|\Theta^{(i)})p(\boldsymbol{x}_j|\boldsymbol{\mu}_k^{(i)}, \boldsymbol{\Sigma}_k^{(i)})}{\sum_{k=1}^{K} P(q_k^{(i)}|\Theta^{(i)})p(\boldsymbol{x}_j|\boldsymbol{\mu}_k^{(i)}, \boldsymbol{\Sigma}_k^{(i)})}
\end{aligned}
$$

        **end**

    **end**

    `/* Maximization step:`                                    `*/`

    **for** $k \leftarrow 1$ *to* $K$ **do**

        **for** $j \leftarrow 1$ *to* $n$ **do**

            `// update the means:`

$$
\boldsymbol{\mu}_k^{i+1} = \frac{\sum_{j=1}^{n} \boldsymbol{x}_j P(q_k^{(i)}|\boldsymbol{x}_j, \Theta^{(i)})}{\sum_{j=1}^{n} P(q_k^{(i)}|\boldsymbol{x}_j, \Theta^{(i)})}
$$

            `// update the variances:`

$$
\boldsymbol{\Sigma}_k^{(i+1)} = \frac{\sum_{j=1}^{n} P(q_k^{(i)}|\boldsymbol{x}_j, \Theta^{(i)})(\boldsymbol{x}_j - \boldsymbol{\mu}_k^{(i+1)})(\boldsymbol{x}_j - \boldsymbol{\mu}_k^{(i+1)})^T}{\sum_{j=1}^{N} P(q_k^{(i)}|\boldsymbol{x}_j, \Theta^{(i)})}
$$

            `// update the priors:`

$$
P(q_k(i+1)|\Theta^{(i+1)}) = \frac{1}{n} \sum_{j=1}^{n} P(q_k^{(i)}|\boldsymbol{x}_j, \Theta^{(i)})
$$

        **end**

    **end**

**until** *the total likelihood increase falls under some desired threshold* ;

return $S$

**Algorithm 3**: Computing EM-based similarity measure

## 6. Redundancy Checking and Generating Summary

Once the sentences are scored the easiest way to create summaries is just to output the topmost $N$ sentences until the required summary length is reached. In that case, we are ignoring other factors: such as redundancy and coherence.

As we know that text summarization clearly entails selecting the most salient information and putting it together in a coherent summary. The answer or summary consists of multiple separately extracted sentences from different documents. Obviously, each of the selected text snippets should individually be important. However, when many of the competing sentences are included in the summary the issue of information overlap between parts of the output comes up and a mechanism for addressing redundancy is needed. Therefore, our summarization systems employ two levels of analysis: first a content level where every sentence is scored according to the features or concepts it covers, and second a textual level, when, before being added to the final output, the sentences deemed to be important are compared to each other and only those that are not too similar to other candidates are included in the final answer or summary. Goldstein, Kantrowitz, Mittal, and Carbonell (1999) observed this in what the authors called "Maximum-Marginal-Relevance (MMR)". Following Hovy et al. (2006) we modeled this by BE overlap between an intermediate summary and a to-be-added candidate summary sentence.

We call this overlap ratio R, where R is between 0 and 1 inclusively. Setting $R = 0.7$ means that a candidate summary sentence, $s$, can be added to an intermediate summary, $S$, if the sentence has a BE overlap ratio less than or equal to 0.7.

## 7. Experimental Evaluation

This section describes the results of experiments conducted using DUC[12] 2007 dataset provided by NIST [13]. Some of the questions these experiments address include:

- How do the different features affect the behavior of the summarizer system?

- Which one of the algorithms (K-means, EM and Local Search) performs better for this particular problem?

We used the main task of DUC 2007 for evaluation. The task was:

*"Given a complex question (topic description) and a collection of relevant documents, the task is to synthesize a fluent, well-organized 250-word summary of the documents that answers the question(s) in the topic."*

The documents of DUC 2007 came from the AQUAINT corpus comprising newswire articles from the Associated Press and New York Times (1998-2000) and Xinhua News Agency (1996-2000). NIST assessors developed topics of interest to them and choose a set of 25 documents relevant (document cluster) to each topic. Each topic and its document cluster were given to 4 different NIST assessors including the developer of the topic. The assessor created a 250-word summary of the document cluster that satisfies the information need expressed in the topic statement. These multiple "reference summaries" are used in the evaluation of summary content.

---

12. http://www-nlpir.nist.gov/projects/duc/
13. National Institute of Standards and Technology

The purpose of our experiments is to study the impact of different features. To accomplish this we generated summaries for the 45 topics of DUC 2007 by each of our seven systems defined as below:

- The **LEX** system generates summaries based on only lexical features (Section 4.2): n-gram (n=1,2,3,4), LCS, WLCS, skip bi-gram, head, head synonym and BE overlap.

- The **LEXSEM** system considers only lexical semantic features (Section 4.3): synonym, hypernym/hyponym, gloss, dependency-based and proximity-based similarity.

- The **SYN** system generates summary based on only syntactic feature (Section 4.1.1).

- The **COS** system generates summary based on the graph-based method (Section 4.5).

- The **SYS1** system considers all the features except the syntactic and semantic features (All features except section 4.1).

- The **SYS2** system considers all the features except the semantic feature (All features except section 4.1.2) and

- The **ALL** system generates summaries taking all the features (Section 4) into account.

## 7.1 Automatic Evaluation

**ROUGE** We carried out automatic evaluation of our summaries using the ROUGE (Lin, 2004) toolkit, which has been widely adopted by DUC for automatic summarization evaluation. *ROUGE* stands for "Recall-Oriented Understudy for Gisting Evaluation". It is a collection of measures that determines the quality of a summary by comparing it to reference summaries created by humans. The measures count the number of overlapping units such as n-gram, word-sequences, and word-pairs between the system-generated summary to be evaluated and the ideal summaries created by humans. The available ROUGE measures are: ROUGE-N (N=1,2,3,4), ROUGE-L, ROUGE-W and ROUGE-S. ROUGE-N is n-gram recall between a candidate summary and a set of reference summaries. ROUGE-L measures the longest common subsequence (LCS) which takes into account sentence level structure similarity naturally and identifies longest co-occurring insequence n-grams automatically. ROUGE-W measures the weighted longest common subsequence (WLCS) providing an improvement to the basic LCS method of computation to credit the sentences having the consecutive matches of words. ROUGE-S is the overlap of skip-bigrams between a candidate summary and a set of reference summaries where skip-bigram is any pair of words in their sentence order allowing for arbitrary gaps. Most of these ROUGE measures have been applied in automatic evaluation of summarization systems and achieved very promising results (Lin, 2004).

For all our systems, we report the widely accepted important metrics: ROUGE-2 and ROUGE-SU. We also present the ROUGE-1 scores since this has never been shown to not correlate with human judgement. All the ROUGE measures were calculated by running

ROUGE-1.5.5 with stemming but no removal of stopwords. ROUGE run-time parameters were set as the same as DUC 2007 evaluation setup. They are:

*ROUGE-1.5.5.pl -2 -1 -u -r 1000 -t 0 -n 4 -w 1.2 -m -l 250 -a*

We also show 95% confidence interval of the important evaluation metrics for our systems to report significance for doing meaningful comparison. We use the ROUGE tool for this purpose. ROUGE uses a randomized method named bootstrap resampling to compute the confidence interval. We used 1000 sampling points in the bootstrap resampling.

We report the evaluation scores of one baseline system (The BASE column) in each of the tables in order to show the level of improvement our systems achieve. The baseline system generates summaries by returning all the leading sentences (up to 250 words) in the $\langle TEXT \rangle$ field of the most recent document(s).

While presenting the results we highlight the top two F-scores and bottom one F-score to indicate significance at a glance.

### 7.1.1 Results and Discussion

**The K-means Learning**   Table 1 shows the ROUGE-1 scores for different combinations of features in the K-means learning. It is noticeable that the K-means performs best for the graph-based cosine similarity feature. Note that including syntactic feature does not improve the score. Also, including syntactic and semantic features increases the score but not by a significant amount. Summaries based on only lexical features give us good ROUGE-1 evaluation.

| Scores | LEX | LEXSEM | SYN | COS | SYS1 | SYS2 | ALL | BASE |
|---|---|---|---|---|---|---|---|---|
| Recall | 0.366 | 0.360 | 0.346 | 0.378 | 0.376 | 0.365 | 0.366 | 0.312 |
| Precision | 0.397 | 0.393 | 0.378 | 0.408 | 0.403 | 0.415 | 0.415 | 0.369 |
| F-score | 0.381 | 0.376 | 0.361 | **0.393** | **0.389** | 0.388 | **0.389** | **0.334** |

Table 1: ROUGE-1 measures in K-means learning

Table 2 shows the ROUGE-2 scores for different combinations of features in the K-means learning. Just like ROUGE-1 graph-based cosine similarity feature performs well here. We get a significant improvement in ROUGE-2 score when we include syntactic feature with all other features. Semantic features do not affect the score much. Lexical Semantic features perform well here.

| Scores | LEX | LEXSEM | SYN | COS | SYS1 | SYS2 | ALL | BASE |
|---|---|---|---|---|---|---|---|---|
| Recall | 0.074 | 0.076 | 0.063 | 0.085 | 0.074 | 0.077 | 0.076 | 0.060 |
| Precision | 0.080 | 0.084 | 0.069 | 0.092 | 0.080 | 0.107 | 0.109 | 0.072 |
| F-score | 0.077 | 0.080 | 0.065 | 0.088 | 0.077 | **0.090** | **0.090** | **0.064** |

Table 2: ROUGE-2 measures in K-means learning

As Table 3 shows: ROUGE-SU scores are the best for all features without syntactic and semantic. Including syntactic/semantic features with other features degrades the scores. Summaries based on only lexical features achieve good scores.

| Scores | LEX | LEXSEM | SYN | COS | SYS1 | SYS2 | ALL | BASE |
|---|---|---|---|---|---|---|---|---|
| Recall | 0.131 | 0.127 | 0.116 | 0.139 | 0.135 | 0.134 | 0.134 | 0.105 |
| Precision | 0.154 | 0.152 | 0.139 | 0.162 | 0.176 | 0.174 | 0.174 | 0.124 |
| F-score | 0.141 | 0.138 | 0.126 | 0.149 | **0.153** | **0.152** | **0.152** | **0.112** |

Table 3: ROUGE-SU measures in K-means learning

Table 4 shows the 95% confidence interval (for F-measures in K-means learning) of the important ROUGE evaluation metrics for all our systems in comparison to the confidence interval of the baseline system. It can be seen that our systems have performed significantly better than the baseline system in most of the cases.

| Systems | ROUGE-1 | ROUGE-2 | ROUGE-SU |
|---|---|---|---|
| Baseline | 0.326680 - 0.342330 | 0.060870 - 0.068840 | 0.108470 - 0.116720 |
| LEX | 0.362976 - 0.400498 | 0.064983 - 0.090981 | 0.128390 - 0.157784 |
| LEXSEM | 0.357154 - 0.395594 | 0.069909 - 0.091376 | 0.126157 - 0.151831 |
| SYN | 0.345512 - 0.377525 | 0.056041 - 0.076337 | 0.116191 - 0.136799 |
| COS | 0.372804 - 0.413440 | 0.075127 - 0.104377 | 0.134971 - 0.164885 |
| SYS1 | 0.367817 - 0.408390 | 0.063284 - 0.095170 | 0.132061 - 0.162509 |
| SYS2 | 0.358237 - 0.400000 | 0.065219 - 0.093733 | 0.123703 - 0.153165 |
| ALL | 0.350756 - 0.404275 | 0.066281 - 0.095393 | 0.124157 - 0.159447 |

Table 4: 95% confidence intervals for K-means system

**The EM learning** Table 5 to Table 7 show different ROUGE measures for the feature combinations in the context of the EM learning. It can be easily noticed that for all these measures we get significant amount of improvement in ROUGE scores when we include syntactic and semantic features along with other features. We get 3-15% improvement over SYS1 in F-score when we include syntactic feature and 2-24% improvement when we include syntactic and semantic features. The cosine similarity measure does not perform as well as it did in the K-means experiments. Summaries considering only the lexical features achieve good results.

Table 8 shows the 95% confidence interval (for F-measures in EM learning) of the important ROUGE evaluation metrics for all our systems in comparison to the confidence interval of the baseline system. We can see that our systems have performed significantly better than the baseline system in most of the cases.

**Local Search Technique** The ROUGE scores based on the feature combinations are given in Table 9 to Table 11. Summaries generated by including all features perform the best scores for all the measures. We get 7-15% improvement over SYS1 in F-score when we include syntactic feature and 8-19% improvement over SYS1 in F-score when we include syntactic and semantic features. In this case also lexical features (LEX) perform well but not better than all features (ALL).

| Scores | LEX | LEXSEM | SYN | COS | SYS1 | SYS2 | ALL | BASE |
|---|---|---|---|---|---|---|---|---|
| Recall | 0.383 | 0.357 | 0.346 | 0.375 | 0.379 | 0.399 | 0.398 | 0.312 |
| Precision | 0.415 | 0.390 | 0.378 | 0.406 | 0.411 | 0.411 | 0.399 | 0.369 |
| F-score | 0.398 | 0.373 | 0.361 | 0.390 | 0.395 | **0.405** | **0.399** | **0.334** |

Table 5: ROUGE-1 measures in EM learning

| Scores | LEX | LEXSEM | SYN | COS | SYS1 | SYS2 | ALL | BASE |
|---|---|---|---|---|---|---|---|---|
| Recall | 0.088 | 0.079 | 0.063 | 0.087 | 0.084 | 0.089 | 0.090 | 0.060 |
| Precision | 0.095 | 0.087 | 0.069 | 0.094 | 0.091 | 0.116 | 0.138 | 0.072 |
| F-score | 0.092 | 0.083 | 0.065 | 0.090 | 0.088 | **0.100** | **0.109** | **0.064** |

Table 6: ROUGE-2 measures in EM learning

| Scores | LEX | LEXSEM | SYN | COS | SYS1 | SYS2 | ALL | BASE |
|---|---|---|---|---|---|---|---|---|
| Recall | 0.145 | 0.128 | 0.116 | 0.138 | 0.143 | 0.145 | 0.143 | 0.105 |
| Precision | 0.171 | 0.153 | 0.139 | 0.162 | 0.167 | 0.186 | 0.185 | 0.124 |
| F-score | 0.157 | 0.139 | 0.126 | 0.149 | 0.154 | **0.163** | **0.161** | **0.112** |

Table 7: ROUGE-SU measures in EM learning

| Systems | ROUGE-1 | ROUGE-2 | ROUGE-SU |
|---|---|---|---|
| Baseline | 0.326680 - 0.342330 | 0.060870 - 0.068840 | 0.108470 - 0.116720 |
| LEX | 0.382874 - 0.416109 | 0.075084 - 0.110454 | 0.144367 - 0.172449 |
| LEXSEM | 0.352610 - 0.395048 | 0.070816 - 0.095856 | 0.125276 - 0.154562 |
| SYN | 0.345512 - 0.377525 | 0.056041 - 0.076337 | 0.115713 - 0.136599 |
| COS | 0.366364 - 0.410020 | 0.076088 - 0.104243 | 0.133251 - 0.164110 |
| SYS1 | 0.378068 - 0.413658 | 0.077480 - 0.099739 | 0.141550 - 0.168759 |
| SYS2 | 0.360319 - 0.414068 | 0.073661 - 0.112157 | 0.130022 - 0.171378 |
| ALL | 0.378177 - 0.412705 | 0.077515 - 0.115231 | 0.141345 - 0.164849 |

Table 8: 95% confidence intervals for EM system

| Scores | LEX | LEXSEM | SYN | COS | SYS1 | SYS2 | ALL | BASE |
|---|---|---|---|---|---|---|---|---|
| Recall | 0.379 | 0.358 | 0.346 | 0.375 | 0.382 | 0.388 | 0.390 | 0.312 |
| Precision | 0.411 | 0.390 | 0.378 | 0.406 | 0.414 | 0.434 | 0.438 | 0.369 |
| F-score | 0.394 | 0.373 | 0.361 | 0.390 | 0.397 | **0.410** | **0.413** | **0.334** |

Table 9: ROUGE-1 measures in local search technique

| Scores | LEX | LEXSEM | SYN | COS | SYS1 | SYS2 | ALL | BASE |
|---|---|---|---|---|---|---|---|---|
| Recall | 0.085 | 0.079 | 0.063 | 0.087 | 0.086 | 0.095 | 0.099 | 0.060 |
| Precision | 0.092 | 0.087 | 0.069 | 0.094 | 0.093 | 0.114 | 0.116 | 0.072 |
| F-score | 0.088 | 0.083 | 0.065 | 0.090 | 0.090 | **0.104** | **0.107** | **0.064** |

Table 10: ROUGE-2 measures in local search technique

| Scores | LEX | LEXSEM | SYN | COS | SYS1 | SYS2 | ALL | BASE |
|---|---|---|---|---|---|---|---|---|
| Recall | 0.143 | 0.128 | 0.116 | 0.138 | 0.145 | 0.148 | 0.150 | 0.105 |
| Precision | 0.168 | 0.153 | 0.139 | 0.162 | 0.170 | 0.195 | 0.196 | 0.124 |
| F-score | 0.155 | 0.139 | 0.126 | 0.149 | 0.157 | **0.169** | **0.170** | **0.112** |

Table 11: ROUGE-SU measures in local search technique

Table 12 shows the 95% confidence interval (for F-measures in local search technique) of the important ROUGE evaluation metrics for all our systems in comparison to the confidence interval of the baseline system. We find that our systems have performed significantly better than the baseline system in most of the cases.

### 7.1.2 Comparison

From the results reported above we can see for all three algorithms our systems clearly outperform the baseline system. Table 13 shows the F-scores of the reported ROUGE measures while Table 14 reports the 95% confidence intervals for the baseline system, the best system in DUC 2007, and our three techniques taking all features (ALL) into consideration. We can see that the method based on local search technique outperforms the other two and the EM algorithm performs better than the K-means algorithm. If we analyze deeply, we find that in all cases but ROUGE-SU with local search the confidence intervals do not overlap with the best DUC 2007 system.

| Systems | ROUGE-1 | ROUGE-2 | ROUGE-SU |
|---------|---------|---------|----------|
| Baseline | 0.326680 - 0.342330 | 0.060870 - 0.068840 | 0.108470 - 0.116720 |
| LEX | 0.380464 - 0.409085 | 0.078002 - 0.100107 | 0.143851 - 0.166648 |
| LEXSEM | 0.353458 - 0.394853 | 0.070845 - 0.096261 | 0.125342 - 0.154729 |
| SYN | 0.345512 - 0.377525 | 0.056041 - 0.076337 | 0.115713 - 0.136599 |
| COS | 0.366364 - 0.410020 | 0.076088 - 0.104243 | 0.133251 - 0.164110 |
| SYS1 | 0.381544 - 0.414534 | 0.079550 - 0.101246 | 0.144551 - 0.170047 |
| SYS2 | 0.370310 - 0.415768 | 0.078760 - 0.114175 | 0.141043 - 0.174575 |
| ALL | 0.384897 - 0.416301 | 0.084181 - 0.114753 | 0.146302 - 0.171736 |

Table 12: 95% confidence intervals for local search system

| Algorithms | ROUGE-1 | ROUGE-2 | ROUGE-SU |
|------------|---------|---------|----------|
| Baseline | 0.334 | 0.064 | 0.112 |
| Best System | 0.438 | 0.122 | 0.174 |
| K-means | 0.389 | 0.089 | 0.152 |
| EM | 0.399 | 0.109 | 0.161 |
| Local Search | 0.413 | 0.107 | 0.170 |

Table 13: ROUGE F-scores for different systems

| Algorithms | ROUGE-1 | ROUGE-2 | ROUGE-SU |
|------------|---------|---------|----------|
| Baseline | 0.326680 - 0.342330 | 0.060870 - 0.068840 | 0.108470 - 0.116720 |
| Best System | 0.431680 - 0.445970 | 0.118000 - 0.127680 | 0.169970 - 0.179390 |
| K-means | 0.350756 - 0.404275 | 0.066281 - 0.095393 | 0.124157 - 0.159447 |
| EM | 0.378177 - 0.412705 | 0.077515 - 0.115231 | 0.141345 - 0.164849 |
| Local Search | 0.384897 - 0.416301 | 0.084181 - 0.114753 | 0.146302 - 0.171736 |

Table 14: 95% confidence intervals for different systems

## 7.2 Manual Evaluation

For a sample of 105 summaries[14] drawn from our different systems' generated summaries we conduct an extensive manual evaluation in order to analyze the effectiveness of our approaches. The manual evaluation comprised a Pyramid-based evaluation of contents and a user evaluation to get the assessment of linguistic quality and overall responsiveness.

### 7.2.1 PYRAMID EVALUATION

In the DUC 2007 main task, 23 topics were selected for the optional community-based pyramid evaluation. Volunteers from 16 different sites created pyramids and annotated the peer summaries for the DUC main task using the given guidelines[15]. 8 sites among them created the pyramids. We used these pyramids to annotate our peer summaries to compute the modified pyramid scores[16]. We used the $DUCView.jar$[17] annotation tool for this purpose. Table 15 to Table 17 show the modified pyramid scores of all our systems for the three algorithms. A baseline system's score is also reported. The peer summaries of the baseline system are generated by returning all the leading sentences (up to 250 words) in the $\langle TEXT \rangle$ field of the most recent document(s). From these results we see that all our systems perform better than the baseline system and inclusion of syntactic and semantic features yields better scores. For all three algorithms we can also notice that the lexical semantic features are the best in terms of modified pyramid scores.

### 7.2.2 USER EVALUATION

10 university graduate students judged the summaries for linguistic quality and overall responsiveness. The given score is an integer between 1 (very poor) and 5 (very good) and is guided by consideration of the following factors: 1. Grammaticality, 2. Non-redundancy, 3. Referential clarity, 4. Focus and 5. Structure and Coherence. They also assigned a content responsiveness score to each of the automatic summaries. The content score is an integer between 1 (very poor) and 5 (very good) and is based on the amount of information in the summary that helps to satisfy the information need expressed in the topic narrative. These measures were used at DUC 2007. Table 18 to Table 20 present the average linguistic quality and overall responsive scores of all our systems for the three algorithms. The same baseline system's scores are given for meaningful comparison. From a closer look at these results, we find that most of our systems perform worse than the baseline system in terms of linguistic quality but achieve good scores in case of overall responsiveness. It is also obvious from the tables that the exclusion of syntactic and semantic features often causes lower scores. On the other hand, lexical and lexical semantic features show good overall responsiveness scores for all three algorithms.

---

14. We have 7 systems for each of the 3 algorithms, cumulatively we have 21 systems. Randomly we chose 5 summaries for each of these 21 systems.
15. http://www1.cs.columbia.edu/ becky/DUC2006/2006-pyramid-guidelines.html
16. This equals the sum of the weights of the Summary Content Units (SCUs) that a peer summary matches, normalized by the weight of an ideally informative summary consisting of the same number of contributors as the peer.
17. http://www1.cs.columbia.edu/ ani/DUC2005/Tool.html

| Systems | Modified Pyramid Scores |
|---------|------------------------|
| Baseline | 0.13874 |
| LEX | 0.44984 |
| LEXSEM | 0.51758 |
| SYN | 0.45762 |
| COS | 0.50368 |
| SYS1 | 0.42872 |
| SYS2 | 0.41666 |
| ALL | 0.49900 |

Table 15: Modified pyramid scores for K-means system

| Systems | Modified Pyramid Scores |
|---------|------------------------|
| Baseline | 0.13874 |
| LEX | 0.51894 |
| LEXSEM | 0.53226 |
| SYN | 0.45058 |
| COS | 0.48484 |
| SYS1 | 0.47758 |
| SYS2 | 0.44734 |
| ALL | 0.49756 |

Table 16: Modified pyramid scores for EM system

| Systems | Modified Pyramid Scores |
|---------|------------------------|
| Baseline | 0.13874 |
| LEX | 0.49760 |
| LEXSEM | 0.53912 |
| SYN | 0.43512 |
| COS | 0.49510 |
| SYS1 | 0.46976 |
| SYS2 | 0.46404 |
| ALL | 0.47944 |

Table 17: Modified pyramid scores for local search system

| Systems | Linguistic Quality | Overall Responsiveness |
|---------|-------------------|------------------------|
| Baseline | 4.24 | 1.80 |
| LEX | 3.08 | 3.20 |
| LEXSEM | 4.08 | 3.80 |
| SYN | 3.24 | 3.60 |
| COS | 4.00 | 3.60 |
| SYS1 | 2.72 | 2.20 |
| SYS2 | 3.12 | 2.80 |
| ALL | 3.56 | 3.80 |

Table 18: Linguistic quality and responsive scores for K-means system

| Systems | Linguistic Quality | Overall Responsiveness |
|---------|-------------------|------------------------|
| Baseline | 4.24 | 1.80 |
| LEX | 4.08 | 4.40 |
| LEXSEM | 3.56 | 3.40 |
| SYN | 4.20 | 3.80 |
| COS | 3.80 | 4.00 |
| SYS1 | 3.68 | 3.80 |
| SYS2 | 4.20 | 3.60 |
| ALL | 3.36 | 3.40 |

Table 19: Linguistic quality and responsive scores for EM system

| Systems | Linguistic Quality | Overall Responsiveness |
|---------|-------------------|------------------------|
| Baseline | 4.24 | 1.80 |
| LEX | 3.24 | 2.40 |
| LEXSEM | 3.12 | 4.20 |
| SYN | 2.64 | 2.00 |
| COS | 3.40 | 3.40 |
| SYS1 | 3.40 | 3.60 |
| SYS2 | 3.12 | 3.80 |
| ALL | 3.20 | 3.20 |

Table 20: Linguistic quality and responsive scores for local search system

## 8. Conclusion and Future Work

In this paper we presented our works on answering complex questions. We extracted eighteen important features for each of the sentences in the document collection. Later we used a simple local search technique to fine-tune the feature weights. For each weight, $w_i$, the algorithm achieves the local maximum of the ROUGE value. In this way, once we learn the weights we rank the sentences by multiplying the feature-vector with the weight-vector. We also experimented with two unsupervised learning techniques: 1) EM and 2) K-means with the features extracted. We assume that we have two clusters of sentences: 1. query-relevant and 2. query-irrelevant. We learned the means of the clusters using the K-means algorithm then we used Bayesian model in order to rank the sentences. The learned means in the K-means algorithm are used as the initial means in the EM algorithm. We applied the EM algorithm to cluster the sentences into two classes : 1) query-relevant and 2) query-irrelevant. We take out the query-relevant sentences and rank them using the learned weights (i.e. in local search). For each of our methods of generating summaries we filter out the redundant sentences using a redundancy checking module and generate summaries by taking the top $N$ sentences.

We also experimented with the effects of different kinds of features. We evaluated our systems automatically using ROUGE and report the significance of our results through 95% confidence intervals. We conducted two types of manual evaluation: 1) Pyramid and 2) User Evaluation to further analyze the performance of our systems. Our experimental results mostly show the following: (a) our approaches achieve promising results, (b) the empirical approach based on a local search technique outperforms the other two learning techniques and EM performs better than the K-means algorithm, (c) our systems achieve better results when we include the tree kernel based syntactic and semantic features, and (d) in all cases but ROUGE-SU with local search the confidence intervals do not overlap with the best DUC 2007 system.

We are now experimenting with the supervised learning techniques (i.e. SVM, MAX-ENT, CRF etc) and analyzing how they perform for this problem. Prior to that, we produced huge amount of labeled data automatically using similarity measures such as ROUGE (Toutanova et al., 2007).

In the future we plan to decompose the complex questions into several simple questions before measuring the similarity between the document sentence and the query sentence. This will certainly serve to create more limited trees and subsequences which might increase the precision. Thus, we expect that by decomposing complex questions into the sets of subquestions that they entail systems can improve the average quality of answers returned and achieve better coverage for the question as a whole.

## Acknowledgments

We thank the anonymous reviewers for their useful comments on the earliest version of this paper. Special thanks go to our colleagues for proofreading the paper. We are also grateful to all the graduate students who took part in the user evaluation process. The research reported here was supported by the Natural Sciences and Engineering Research Council (NSERC) research grant and the University of Lethbridge.

# Appendix A. Stop Word List

| | | | | | |
|---|---|---|---|---|---|
| reuters | ap | jan | feb | mar | apr |
| may | jun | jul | aug | sep | oct |
| nov | dec | tech | news | index | mon |
| tue | wed | thu | fri | sat | 's |
| a | a's | able | about | above | according |
| accordingly | across | actually | after | afterwards | again |
| against | ain't | all | allow | allows | almost |
| alone | along | already | also | although | always |
| am | amid | among | amongst | an | and |
| another | any | anybody | anyhow | anyone | anything |
| anyway | anyways | anywhere | apart | appear | appreciate |
| appropriate | are | aren't | around | as | aside |
| ask | asking | associated | at | available | away |
| awfully | b | be | became | because | become |
| becomes | becoming | been | before | beforehand | behind |
| being | believe | below | beside | besides | best |
| better | between | beyond | both | brief | but |
| by | c | c'mon | c's | came | can |
| can't | cannot | cant | cause | causes | certain |
| certainly | changes | clearly | co | com | come |
| comes | concerning | consequently | consider | considering | contain |
| containing | contains | corresponding | could | couldn't | course |
| currently | d | definitely | described | despite | did |
| didn't | different | do | does | doesn't | doing |
| don't | done | down | downwards | during | e |
| each | edu | eg | e.g. | eight | either |
| else | elsewhere | enough | entirely | especially | et |
| etc | etc. | even | ever | every | everybody |
| everyone | everything | everywhere | ex | exactly | example |
| except | f | far | few | fifth | five |
| followed | following | follows | for | former | formerly |
| forth | four | from | further | furthermore | g |
| get | gets | getting | given | gives | go |
| goes | going | gone | got | gotten | greetings |
| h | had | hadn't | happens | hardly | has |
| hasn't | have | haven't | having | he | he's |
| hello | help | hence | her | here | here's |
| hereafter | hereby | herein | hereupon | hers | herself |
| hi | him | himself | his | hither | hopefully |
| how | howbeit | however | i | i'd | i'll |
| i'm | i've | ie | i.e. | if | ignored |
| immediate | in | inasmuch | inc | indeed | indicate |
| indicated | indicates | inner | insofar | instead | into |
| inward | is | isn't | it | it'd | it'll |
| it's | its | itself | j | just | k |
| keep | keeps | kept | know | knows | known |
| l | lately | later | latter | latterly | least |
| less | lest | let | let's | like | liked |
| likely | little | look | looking | looks | ltd |
| m | mainly | many | may | maybe | me |
| mean | meanwhile | merely | might | more | moreover |
| most | mostly | mr. | ms. | much | must |
| my | myself | n | namely | nd | near |
| nearly | necessary | need | needs | neither | never |
| nevertheless | new | next | nine | no | nobody |
| non | none | noone | nor | normally | not |
| nothing | novel | now | nowhere | o | obviously |
| of | off | often | oh | ok | okay |
| old | on | once | one | ones | only |
| onto | or | other | others | otherwise | ought |
| our | ours | ourselves | out | outside | over |
| overall | own | p | particular | particularly | per |
| perhaps | placed | please | plus | possible | presumably |
| probably | provides | q | que | quite | qv |
| r | rather | rd | re | really | reasonably |
| regarding | regardless | regards | relatively | respectively | right |
| s | said | same | saw | say | saying |
| says | second | secondly | see | seeing | seem |
| seemed | seeming | seems | seen | self | selves |
| sensible | sent | serious | seriously | seven | several |
| shall | she | should | shouldn't | since | six |
| so | some | somebody | somehow | someone | something |
| sometime | sometimes | somewhat | somewhere | soon | sorry |
| specified | specify | specifying | still | sub | such |
| sup | sure | t | t's | take | taken |
| tell | tends | th | than | thank | thanks |
| thanx | that | that's | thats | the | their |
| theirs | them | themselves | then | thence | there |
| there's | thereafter | thereby | therefore | therein | theres |
| thereupon | these | they | they'd | they'll | they're |
| they've | think | third | this | thorough | thoroughly |
| those | though | three | through | throughout | thru |
| thus | to | together | too | took | toward |
| towards | tried | tries | truly | try | trying |
| twice | two | u | un | under | unfortunately |
| unless | unlikely | until | unto | up | upon |
| us | use | used | useful | uses | using |
| usually | uucp | v | value | various | very |
| via | viz | vs | w | want | wants |
| was | wasn't | way | we | we'd | we'll |
| we're | we've | welcome | well | went | were |
| weren't | what | what's | whatever | when | whence |
| whenever | where | where's | whereafter | whereas | whereby |
| wherein | whereupon | wherever | whether | which | while |
| whither | who | who's | whoever | whole | whom |
| whose | why | will | willing | wish | with |
| within | without | won't | wonder | would | would |
| wouldn't | x | y | yes | yet | you |
| you'd | you'll | you're | you've | your | yours |
| yourself | yourselves | z | zero | | |